\PassOptionsToPackage{table}{xcolor}
\documentclass[sigconf,screen]{acmart}
\renewcommand\footnotetextcopyrightpermission[1]{} 
\AtBeginDocument{%
  \providecommand\BibTeX{{%
    Bib\TeX}}}

\acmConference[ACM MM '26]{The 34th ACM International Conference on Multimedia}{November 10--14, 2026}{Rio de Janeiro, Brazil}
\usepackage{xcolor}
\usepackage{pifont}
\newcommand{\cmark}{\ding{51}}
\newcommand{\xmark}{\ding{55}}
\usepackage{multirow}
\usepackage{booktabs}
\usepackage{graphicx} 
\usepackage{booktabs} 
\usepackage{makecell}
\usepackage{tabularx}
\usepackage{array}
\usepackage{graphicx}
\usepackage{subcaption}
\usepackage[most]{tcolorbox}
\usepackage{listings}

\newtcblisting{systemprompt}[1][]{%
  enhanced,
  breakable,
  listing only,
  listing options={
    basicstyle=\ttfamily\scriptsize,
    breaklines=true,
    breakatwhitespace=false,
    columns=fullflexible,
    keepspaces=true,
    showstringspaces=false,
  },
  colback=blue!5,
  colframe=blue!55!black,
  colbacktitle=blue!55!black,
  coltitle=white,
  title=System Instruction,
  boxrule=0.5pt,
  arc=2mm,
  left=8pt,
  right=8pt,
  top=8pt,
  bottom=8pt,
  fonttitle=\bfseries,
  #1
}

\usepackage{fancyhdr}

\fancypagestyle{plain}{%
  \fancyhf{}%
  \fancyfoot[C]{\thepage}%
  \renewcommand{\headrulewidth}{0pt}%
  \renewcommand{\footrulewidth}{0pt}%
}
\begin{document}

\title{MTAVG-Bench 2.0: Diagnosing Failure Modes of Cinematic Expressiveness in Multi-Talker Audio-Video Generation}

\author{%
{\large Haitian Li\textsuperscript{1,4,\textdagger}, Yanghao Zhou\textsuperscript{2,\textdagger,\textdaggerdbl}, Heyan Huang\textsuperscript{2}, Liangji Chen\textsuperscript{3,1}, YiMing Cheng\textsuperscript{4}, Xu Liu\textsuperscript{5},}\\
{\large Dian Jin\textsuperscript{5}, Jiajun Xu\textsuperscript{6}, Jingyun Liao\textsuperscript{6}, Tian Lan\textsuperscript{2}, Ziqin Zhou\textsuperscript{7}, Yueying Liu\textsuperscript{8},}\\
{\large Yu Bai\textsuperscript{9}, Changsen Yuan\textsuperscript{8,*}, Jinxing Zhou\textsuperscript{10}, Xian-Ling Mao\textsuperscript{2}, Xuefeng Chen\textsuperscript{6}, Yousheng Feng\textsuperscript{6}}\\[0.55em]
{\normalsize \textsuperscript{1}Shanghai University \quad
\textsuperscript{2}Beijing Institute of Technology \quad
\textsuperscript{3}Shanghai Film Academy \quad
\textsuperscript{4}Tsinghua University}\\
{\normalsize \textsuperscript{5}Hefei University of Technology \quad
\textsuperscript{6}Inkeverse Group Limited \quad
\textsuperscript{7}The University of Adelaide \quad
\textsuperscript{8}Beijing University of Technology}\\
{\normalsize \textsuperscript{9}Beijing Academy of Artificial Intelligence \quad
\textsuperscript{10}OpenNLP Lab}\\[0.25em]
{\normalsize \textsuperscript{\textdagger}Equal contribution. \quad \textsuperscript{\textdaggerdbl}Project Leader. \quad \textsuperscript{*}Corresponding author.}%
}
\renewcommand{\shortauthors}{Li et al.}

\begin{abstract}
In recent years, Multi-Talker Audio-Video Generation (MTAVG) models have shown promising performance on fundamental metrics such as lip-sync and audio-visual alignment. However, these metrics remain insufficient for assessing cinematic expressiveness in scene-level generation. In multi-character scenes, generation models must go beyond audio-visual realism to convey coherent character performance and other higher-level cinematic qualities. To fill this gap, we introduce MTAVG-Bench 2.0, a benchmark for diagnosing failure modes of cinematic expressiveness in multi-talker audio-video generation. Unlike prior settings that mainly focus on the quality of basic multi-turn dialogue, MTAVG-Bench 2.0 targets short-drama and scene-level generation, and establishes a high-level failure taxonomy spanning acting, narrative, atmosphere, and audio-visual language. Based on this taxonomy, we construct more than 10,000 question-answering evaluation instances, together with subsets for short-drama-level assessment and temporal localization of failure modes, to systematically evaluate the ability of omni large language models to diagnose high-level audio-visual failures. Experimental results show that commercial omni models such as Gemini substantially outperform other evaluators, yet even the strongest models continue to struggle with complex failures in our benchmark. These results demonstrate that MTAVG-Bench 2.0 provides a systematic benchmark for failure diagnosis in cinematic multi-talker audio-video generation.
\end{abstract}

\maketitle\pagestyle{plain}
\thispagestyle{plain}

\begin{center}
\small
\textbf{GitHub:} \url{https://github.com/ChinChilla-HTL/MTAVG-Bench2} \quad
\textbf{Hugging Face:} \url{https://huggingface.co/datasets/Lanht/MTAVG-Bench2}
\end{center}




\section{Introduction}

In recent years, audio-video generation has advanced substantially~\cite{hacohen2026ltx,low2025ovi,team2026mova,guo2026dreamid}. In particular, for multi-speaker dialogue scenarios, existing models are increasingly capable of generating multi-character videos with a strong sense of realism, achieving a relatively high level of performance in speech synchronization, motion naturalness, and audio-visual alignment. Benefiting from continued improvements in multimodal modeling~\cite{xu2025qwen25omnitechnicalreport,cheng2024videollama,team2023gemini,liu2025ola,ye2025omnivinci,yao2024minicpmvgpt4vlevelmllm,ai2025ming,dai2025see} and audio-visual generation, an increasing number of approaches are evolving from standalone video generation toward joint audio-video generation, while also targeting more complex generation objectives such as character interaction, emotional expression, and scene-level storytelling~\cite{zhang2025stage,zhang2025evaluation}. In this context, multi-speaker dialogue scenes, which inherently involve character relationships, information exchange, and emotional communication, are becoming an increasingly important setting for multimodal scene generation and evaluation~\cite{shi2026msvbench,han2025video,mao2024tavgbench,zhou2026avgen}.

However, realistic multi-speaker scene generation cannot be fully characterized by low-level fidelity alone. In more cinematic settings, such as short dramas, a successful scene requires not only local correctness in lip synchronization, audio quality, and basic audio-visual alignment, but also expressive character performance, believable interpersonal dynamics, a coherent emotional atmosphere, and visual organization that supports dramatic progression. By contrast, existing evaluation protocols for multi-speaker audio-video generation still focus primarily on whether basic audio-visual capabilities are correct, such as speech quality and naturalness, audio-visual alignment, and the synchronization of lip movements. Although these criteria are necessary for assessing generation quality, they are often insufficient to determine whether a clip can truly function as a cinematic scene~\cite{hua2025vabench,cao2025t2av,liu2025javisgpt}. They also provide limited support for diagnosing high-level failures, such as flat performance, weak emotional progression, an unconvincing atmosphere, or shot organization that lacks narrative motivation. In other words, current evaluation mainly addresses whether a clip is reasonable at the dialogue level. For cinematic scene generation, we must further ask whether the scene is sufficiently expressive, whether it establishes a convincing atmosphere, and whether it exhibits coherent cinematic organization over time.

\begin{figure*}
    \centering
    \includegraphics[width=1\linewidth]{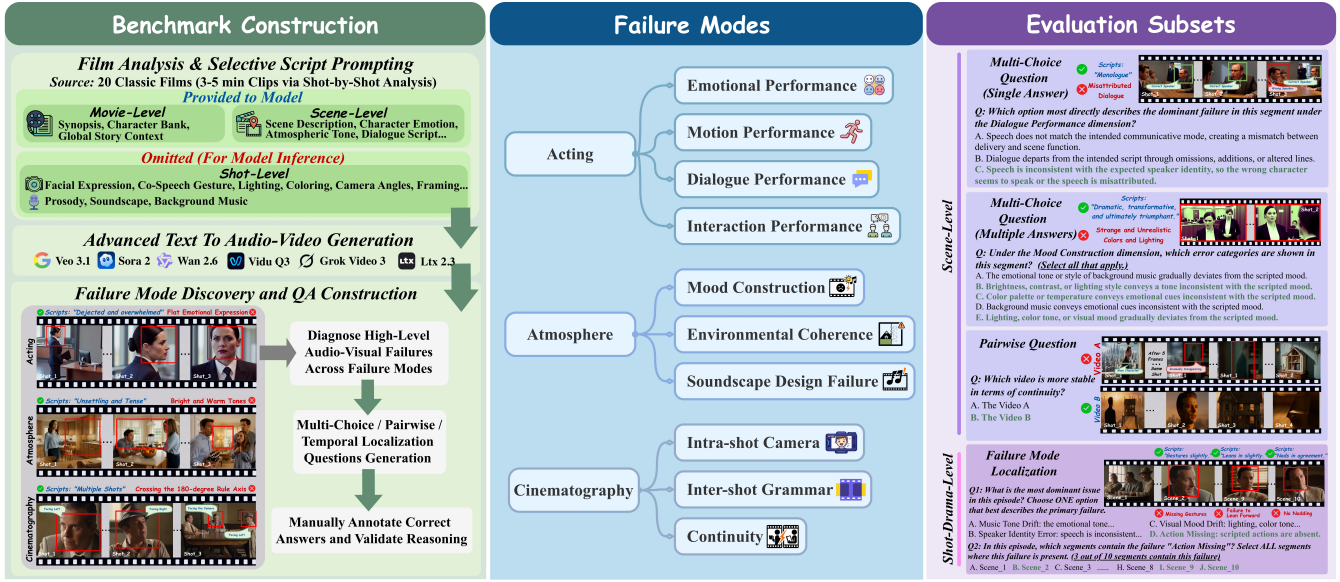}
    \caption{\textbf{Overview of the MTAVG-Bench 2.0 framework.} 
    The benchmark is constructed through a pipeline of film analysis and selective script prompting, text-to-audio-video generation, failure discovery, question generation, and manual validation. 
    }
    \label{fig:figure_main}
\end{figure*}

For scene-level generation, merely determining whether a clip is overall \textit{acceptable} fails to capture the intricacies of cinematic requirements. To provide meaningful insights, scene-level assessment demands a multi-dimensional approach where an evaluator can go beyond holistic scoring to identify specific types of failures, marking them in an interpretable and temporally grounded manner.
Compared with a simple quality judgment, such structured diagnosis not only helps reveal the limitations of current generation systems, but also provides necessary support for advancing toward more expressive and narratively coherent audio-visual generation. To fill this gap, we introduce \textbf{MTAVG-Bench 2.0}, a benchmark for scene-level diagnosis of high-level failure modes in multi-talker audio-video generation. The benchmark focuses on short-drama and scene-level generation, and organizes failures into a three-dimensional taxonomy covering \textbf{acting}, \textbf{atmosphere}, and \textbf{cinematography}. We choose these three dimensions not because they exhaust all aspects of scene quality, but because they provide a compact, operational, and diagnostically meaningful decomposition of cinematic expressiveness. Specifically, \textbf{acting} concerns character performance and interpersonal interaction, \textbf{atmosphere} concerns affective construction and emotional tone, and \textbf{cinematography} concerns the audio-visual organization and shot presentation of a scene as it unfolds over time.

As illustrated in Table.~\ref{tab:paradigm_compare}. Based on this taxonomy, we construct more than 10,000 question-answering instances, and further design dedicated subsets for \textbf{scene-level assessment} and \textbf{temporal localization of failure modes}, thereby operationalizing high-level cinematic evaluation as a structured diagnosis task rather than a binary quality judgment, as shown in Figure~\ref{fig:figure_main}. Using MTAVG-Bench 2.0, we systematically evaluate the diagnostic ability of current mainstream omni audio-visual understanding models. Experimental results show that, although leading commercial omni models achieve the strongest overall performance, even the best current evaluator still shows clear limitations when handling complex cases involving character performance. This suggests that high-level failure diagnosis for cinematic multi-talker generation remains far from solved, and further indicates that evaluating scene-level expressiveness requires substantially richer benchmarks than those based solely on conventional fidelity metrics.

\begin{figure*}
    \centering
    \includegraphics[width=1\linewidth]{figures/pipeline.jpg}
    \caption{\textbf{Overview of MTAVG-Bench 2.0 construction framework.} The pipeline consists of three stages. First, classical film scenes are decomposed into hierarchical script prompts, which are used to generate candidate videos with advanced text-to-audio-video models. Second, annotators identify high-level failure evidence and map each case to taxonomy-defined failure modes. Third, failure-aware diagnostic QA pairs are constructed from the reviewed evidence and further refined through human annotation, expert discussion, and verification, resulting in diverse evaluation instances for fine-grained failure diagnosis.}
    \label{fig:data_pipeline}
\end{figure*}

\textbf{In summary, our contributions are three-fold:}

\begin{itemize}
\item We formulate high-level failure diagnosis as a distinct evaluation problem for scene-level cinematic expressiveness in multi-talker audio-video generation, moving beyond conventional evaluation centered on low-level fidelity and local interaction quality.
\item We introduce MTAVG-Bench 2.0, a benchmark tailored to short-drama and scene-level generation, with a compact taxonomy spanning acting, atmosphere, and cinematography. Based on this taxonomy, we construct over 10,000 evaluation instances, together with dedicated subsets for scene-level assessment and temporal localization, enabling fine-grained and temporally grounded diagnosis of high-level failures.
\item We systematically benchmark contemporary omni audio-visual understanding models on this setting and show that, although leading commercial systems perform best overall, diagnosing complex failures in character performance remains challenging even for the strongest evaluators, highlighting the need for richer benchmarks.
\end{itemize}

\section{Related Work}

\subsection{Cinematic Video Generation}
Cinematic video generation is evolving from short-clip text-to-video synthesis toward more structured text-to-audio-video generation, where the objective extends beyond visual realism to long-range narrative coherence, scene continuity, controllable cinematic expression, and coordinated audiovisual dynamics. Compared with generic video generation, this setting requires models to reason over scene transitions, story structure, camera language, and character interaction across extended temporal contexts. To address these challenges, recent approaches increasingly rely on structured intermediate representations, such as scripts, storyboards, keyframes, or long-context scene abstractions, to guide multi-scene composition and improve temporal consistency~\cite{lin2023videodirectorgpt,zhu2023moviefactory,xiao2025captain,guo2025long,shi2025pvchat}. In parallel, cinematic generation is moving toward tighter integration of filmmaking priors into the synthesis process, incorporating camera rules, differentiable filming, retrieval-guided cinematography, and rhythm-aware control to better model cinematic language and production dynamics~\cite{rao2023dynamic,jiang2024cinematic,huang2025filmaster}. More recently, generation pipelines have become increasingly structured and production-oriented, with staged formulations that explicitly connect planning, directing, and evaluation or control to executable video outputs~\cite{mu2026scriptagent,yangshotverse}.

At the same time, audiovisual generation is becoming an essential component of film-grade systems. JavisDiT~\cite{liu2025javisdit,liu2026javisdit++,liu2025javisgpt} and the dense cinematic caption supervision adopted in Seedance 1.0~\cite{gao2025seedance} further extend this trend by improving audio-video synchrony and enriching the representation of scenes, actions, and camera semantics. Taken together, these developments indicate a broader transition from appearance-driven video synthesis to unified modeling of narrative structure, cinematic control, and multimodal consistency. They also make evaluation increasingly challenging, especially for dialogue-centric cinematic videos involving multiple speakers, where generation quality depends not only on visual plausibility and audiovisual alignment, but also on speaker-aware interaction, dialogue coherence, and high-level cinematic expressiveness. This motivates a benchmark specifically designed for evaluating multi-talker dialogue cinematic video generation.

\subsection{Comprehensive Benchmark for Audio-video Generation}
As audio-video generation progresses from short clips toward more structured and cinematic scenarios, evaluation has expanded beyond generic perceptual quality to cover semantic alignment, multimodal consistency, and task-specific failure analysis. Early and recent benchmarks such as Harmony-Bench~\cite{hu2025harmony}, JavisBench~\cite{liu2025javisdit}, UniAVGen~\cite{zhang2025uniavgen}, VerseBench~\cite{wang2025universe}, and VABench~\cite{hua2025vabench} mainly assess prompt following, audio-video correspondence, and perceptual quality under text-to-audio-video or image-to-audio-video settings. These benchmarks provide important standardized testbeds for synchronized generation, but their emphasis is still largely on low- to mid-level quality and alignment, rather than on higher-level cinematic structure. In parallel, benchmarks such as MovieBench~\cite{wu2025moviebench} and CineTechBench~\cite{wang2025cinetechbench} move evaluation closer to movie-level generation by explicitly considering hierarchical narrative organization and cinematographic technique, thereby reflecting the growing need to assess long-range coherence and film-oriented controllability.

Another line of work focuses more explicitly on diagnostic evaluation. PhyAVBench~\cite{xie2025phyavbench} examines physical consistency in generated audio-video outputs, while VideoHallu~\cite{li2025videohallu} and Pistachio~\cite{li2025pistachio} introduce failure-oriented settings that probe hallucination and fine-grained error categories. Despite this progress, existing benchmarks remain fragmented: some emphasize perceptual realism and alignment, some target specific failure modes, and others focus on narrative or cinematographic control. Relatively few benchmarks jointly evaluate speaker-centric generation, multi-speaker interaction, dialogue structure, and cinematic expressiveness within a unified framework. This gap motivates benchmarks that support both broad multimodal coverage and interpretable diagnosis of high-level cinematic failures in complex audio-visual generation settings.

\begin{table*}[t]
\centering
\caption{Comparison of evaluation paradigms. Existing benchmarks mainly assess perceptual quality and alignment, while MTAVG-Bench 2.0 further introduces a high-level cinematic taxonomy spanning acting, atmosphere, and cinematography, together with multi-speaker dialogue structure and failure diagnosis.}
\resizebox{\textwidth}{!}{
\begin{tabular}{lccccccccc}
\toprule
\textbf{Benchmarks} & 
\textbf{\#Video} & 
\textbf{\#QA} & 
\textbf{Dimen.} & 
\textbf{Failure-Mode} & 
\textbf{Modalities} &  
\textbf{Speaker-Centric}  &
\textbf{Multi-speaker}  &
\textbf{Dialogue} &
\textbf{Cine. Express.} \\
\midrule

Harmony-Bench~\cite{hu2025harmony} 
& 150 & -- & 3 & -- & T2AV & \xmark & \xmark & \xmark & \xmark \\

JavisBench~\cite{liu2025javisdit}& 10{,}140 & -- & 5 & -- & T2AV & \xmark & \xmark & \xmark & \xmark \\

UniAVGen~\cite{zhang2025uniavgen}      
& 100 & -- & 3 & -- & T2AV & \xmark & \xmark & \xmark & \xmark \\

VerseBench~\cite{wang2025universe}    
& 600 & -- & 4 & -- & T2AV & \xmark & \xmark & \xmark & \xmark \\

VABench~\cite{hua2025vabench}       
& 1,300 & 14,300 & 15 & -- & T2AV/I2AV & \xmark & \xmark & \xmark & \xmark \\


PhyAVBench~\cite{xie2025phyavbench}
& 20,000 & -- & 6 & -- & T2AV & \xmark & \xmark & \xmark & \xmark \\

VideoHallu~\cite{li2025videohallu} 
& 120 & 3,233 & 4 & 13 & T2V & \xmark & \xmark & \xmark & \xmark \\

Pistachio~\cite{li2025pistachio}
& 4,962 & -- & 5 & 31 & T2V & \xmark & \xmark & \xmark & \xmark \\

MTAVG-Bench ~\cite{zhou2026mtavg}
& 1,880 
& 2,410 
& 9 
& 37 
& T2AV 
& \cmark 
& \cmark 
& \cmark 
& \xmark \\

\midrule

\rowcolor{lightgray}
MTAVG-Bench 2.0 (Ours) 
& 2,466& 11,600& 10& 45& T2AV/I2AV& \cmark 
& \cmark 
& \cmark 
& \cmark \\

\bottomrule
\end{tabular}
}

\label{tab:paradigm_compare}

\end{table*}

\begin{table*}[t]
\centering
\small
\setlength{\tabcolsep}{4pt}
\renewcommand{\arraystretch}{1.18}
\caption{Hierarchical taxonomy of high-level failure modes in MTAVG-Bench. The taxonomy is organized into three major categories---acting, atmosphere, and cinematography---with ten sub-dimensions covering scene-level cinematic expressiveness.}
\begin{tabularx}{\textwidth}{
>{\raggedright\arraybackslash\bfseries}p{0.19\textwidth}
>{\raggedright\arraybackslash\bfseries}p{0.28\textwidth}
>{\raggedright\arraybackslash}X
}
\toprule
\textbf{Major Categories} & \textbf{Sub-dimensions} & \textbf{Evaluation Focus} \\
\midrule

\multirow{4}{*}{\textbf{Acting}}
& Emotional Performance (EP)
& Emotional expressiveness of characters, including facial expression, gesture, vocal affect, and multimodal consistency with the scripted emotional state. \\
\cline{2-3}
& Motion Performance (MP)
& Naturalness and script consistency of character motion, including action execution, motion dynamics, and intention continuity. \\
\cline{2-3}
& Dialogue Performance (DP)
& Correctness of spoken delivery, including speech mode, dialogue content, and speaker identity consistency. \\
\cline{2-3}
& Interaction Performance (IP)
& Plausibility of inter-character interaction, including gaze coordination, turn-taking, spatial relationship, targeting, and environmental responsiveness. \\
\midrule

\multirow{3}{*}{\textbf{Atmosphere}}
& Mood Construction (MC)
& Consistency between the intended scene mood and the conveyed affective cues from lighting, color, music, and overall audio-visual tone. \\
\cline{2-3}
& Environmental Coherence (EC)
& Realism and temporal coherence of the environment, including scene plausibility, spatial depth, and smoothness of scene transitions. \\
\cline{2-3}
& Soundscape Design (SD)
& Quality and appropriateness of non-speech audio, including ambient sound presence, background-audio balance, and scripted sound events. \\
\midrule

\multirow{3}{*}{\textbf{Cinematography}}
& Intra-shot Camera (IC)
& Appropriateness of camera behavior within a shot, including framing, motion, focus, blocking, and alignment with narrative attention. \\
\cline{2-3}
& Inter-shot Grammar (IG)
& Logical progression of shot composition across adjacent shots, including shot sequencing and adherence to cinematic grammar rules. \\
\cline{2-3}
& Continuity (CT)
& Preservation of action, spatial, and character continuity across shots, ensuring coherent scene progression and visual consistency. \\
\bottomrule
\end{tabularx}
\label{tab:taxonomy}
\end{table*}

\section{MTAVG-Bench 2.0}

In this section, we present MTAVG-Bench 2.0, a benchmark designed to diagnose high-level cinematic failures in multi-talker audio-video generation. As illustrated in Figure~\ref{fig:data_pipeline}, the benchmark is constructed via a three-stage pipeline, including selective prompt construction from curated film scenes, failure discovery with taxonomy-based mapping on generated videos, and the construction of failure-diagnosis QA data with expert refinement and human verification. In contrast to conventional evaluation protocols that primarily emphasize low-level audio-visual fidelity, MTAVG-Bench 2.0 focuses on scene-level cinematic expressiveness, evaluating whether generated dialogue scenes exhibit believable acting, coherent atmosphere, and appropriate shot organization. By defining a high-level failure taxonomy, collecting human-validated failure cases, and transforming them into structured question-answering instances, MTAVG-Bench 2.0 enables a more comprehensive assessment: not only determining whether a model can recognize problematic videos, but also evaluating its capability to identify specific categories of cinematic failures and, in certain settings, to localize their temporal occurrence.

\begin{figure}[t]
    \centering
    \begin{subfigure}{\linewidth}
        \centering
        \includegraphics[width=\linewidth]{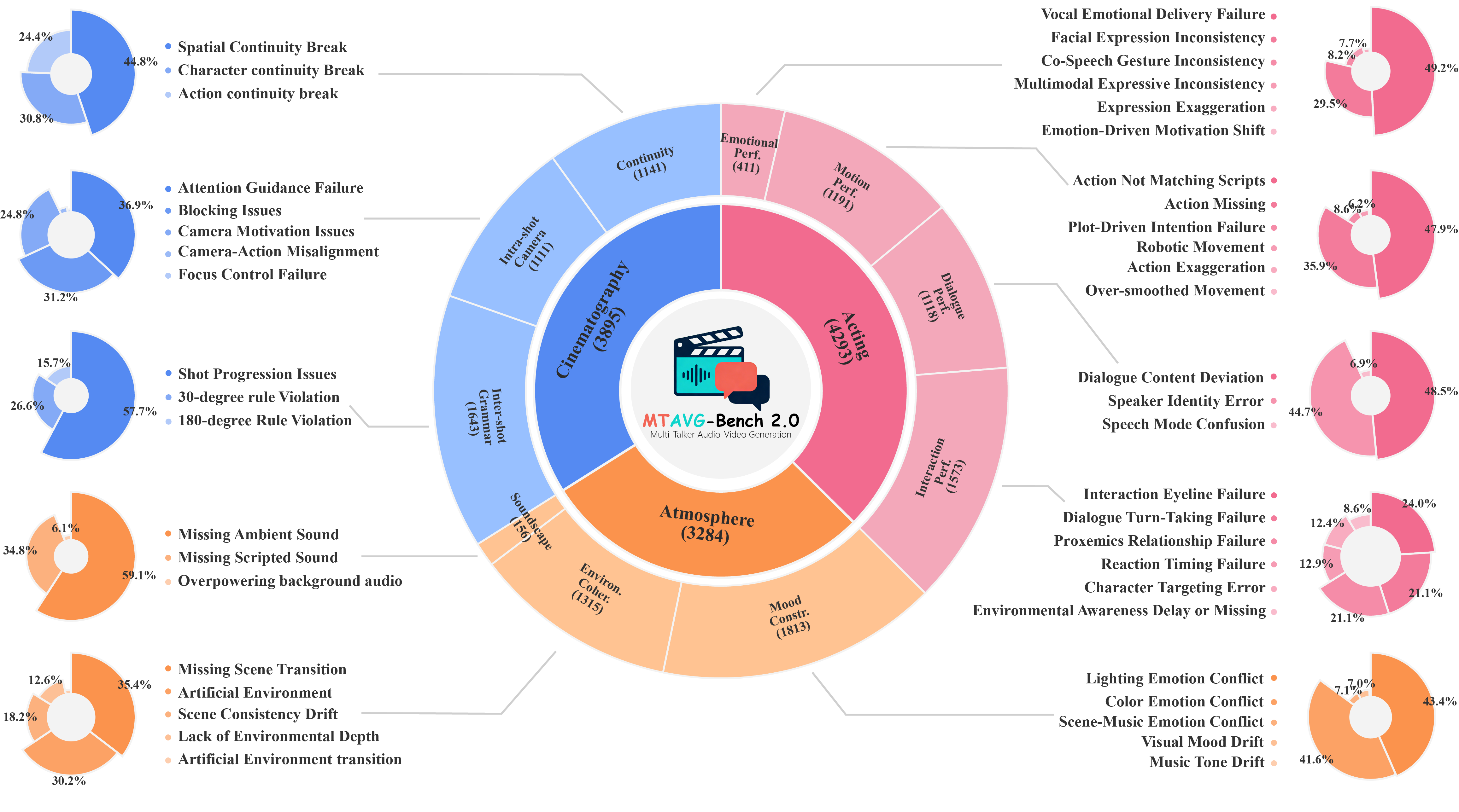}
        \caption{Distribution of failure modes in MTAVG-Bench 2.0.}
        \label{fig:distribution}
    \end{subfigure}

    \begin{subfigure}{\linewidth}
        \centering
        \includegraphics[width=\linewidth]{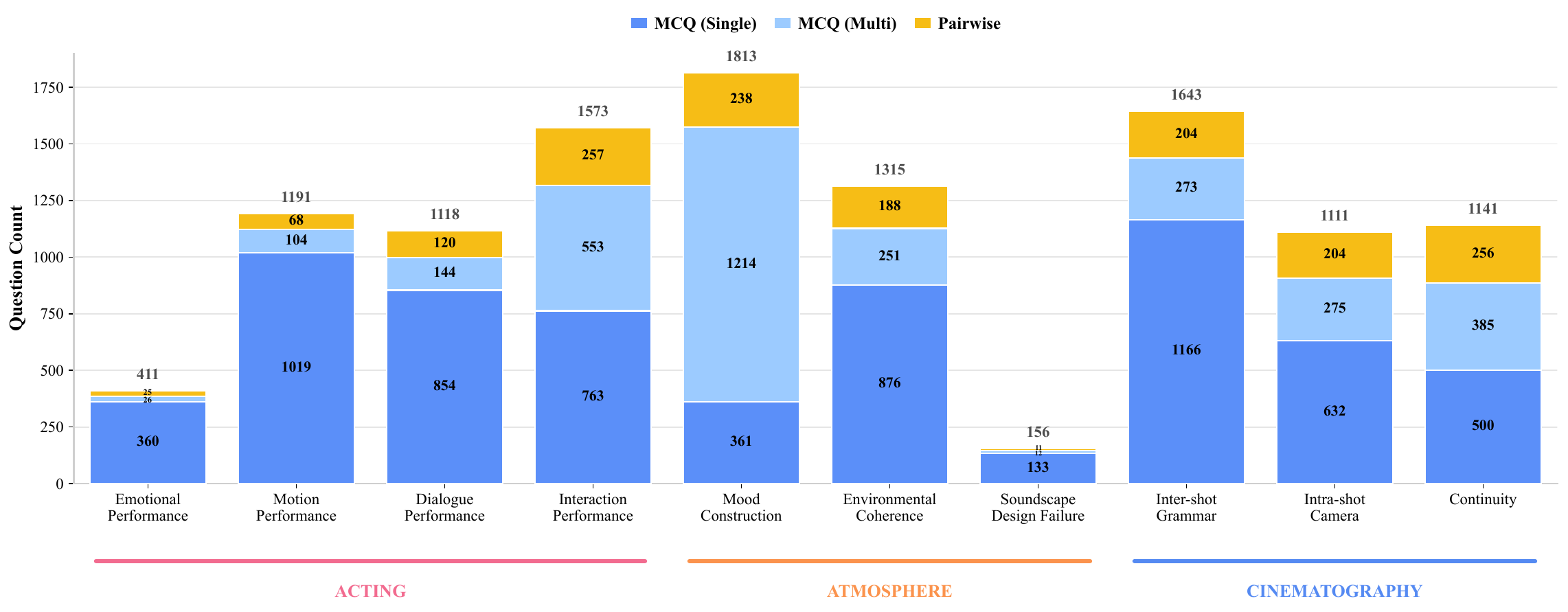}
        \caption{Question-type distribution across sub-dimensions.}
        \label{fig:question_type}
    \end{subfigure}
    
    \caption{\textbf{Dataset statistics of MTAVG-Bench 2.0.} 
    Top: distribution of failure cases across the three major categories and their fine-grained sub-dimensions. 
    Bottom: distribution of question formats across sub-dimensions, including MCQ (single answer), MCQ (multiple answers), and pairwise questions in scene-level questions.}
    \label{fig:dataset_statistics}
\end{figure}

\subsection{Failure Taxonomy}

The core objectives of MTAVG-Bench 2.0 are to extend the failure taxonomy beyond perceptual-level audio-visual breakdowns in multi-character dialogue scenes toward a higher-level, cinema-oriented evaluation framework for short-form narrative videos. To this end, we organize high-level failure modes into three major categories—\textbf{Acting}, \textbf{Atmosphere}, and \textbf{Cinematography}—which characterize scene-level error types in audio-visual generation, including character performance, affective environment, and cinematic organization, as shown in Figure~\ref{fig:dataset_statistics}. Specifically, \textbf{Acting} encompasses failures in emotional expression, motion, dialogue delivery, and interpersonal interaction, focusing on whether characters behave, speak, and respond in a manner consistent with the intended narrative. \textbf{Atmosphere} evaluates whether a scene conveys an appropriate overall mood through coherent audio-visual cues, including mood construction, environmental consistency, and soundscape design. \textbf{Cinematography} assesses whether the audio-visual realization of shooting and editing exhibits coherence with cinematic language, including camera behavior, shot composition, and editing continuity. As summarized in Table~\ref{tab:taxonomy}, within each sub-dimension we further define fine-grained failure modes corresponding to recurring, observable error patterns in generated videos. We also distinguish between failure cases, i.e., concrete problematic phenomena observed in a clip, and failure modes, i.e., abstract taxonomy labels used to categorize such phenomena. The taxonomy is designed to be cinematically grounded, annotation-operational, and evaluation-oriented, thereby enabling fine-grained and temporally localized diagnosis of audio-visual failures in generated multi-talker videos.

\subsection{Benchmark Construction}

Based on the above taxonomy, we construct MTAVG-Bench 2.0 through a multi-stage pipeline, starting from curated cinematic sources and progressively generating and filtering human-reviewed failure annotations. We first collect classic film scenes covering diverse multi-character dialogue settings, emotional relationships, and cinematic staging patterns, and convert them into selective hierarchical prompts at both the short-drama and scene levels. These prompts retain key elements relevant to audio-visual short-form narratives, including plot summaries, character information,  character Emotion, scripted actions, dialogue content, scene descriptions, and atmospheric Tone. Using these prompts, we generate multi-talker audio-visual clips with multiple image+text-to-audio-video systems, preserving sufficient narrative and interactional context to support scene-level diagnosis. Human annotators then review the generated videos to identify observable failure evidence, which is subsequently mapped into the predefined taxonomy via bucket-level failure categorization.

As illustrated in Figure~\ref{fig:data_pipeline}, to ensure benchmark quality, we further perform failure-conditioned data selection, retaining only clips and temporal segments that exhibit clear, verifiable, and diagnostically meaningful failures. The selected samples are then refined through human annotation, with agent-assisted filtering used only to organize candidates and remove obviously low-quality annotations. Following this pipeline, MTAVG-Bench 2.0 comprises 2,466 videos, 10 sub-dimensions, 45 failure modes, and approximately 11.6K QA instances, thereby extending dialogue-centric audio-visual evaluation to higher-level cinematic diagnosis.

\begin{table*}[h]
\centering
\footnotesize
\setlength{\tabcolsep}{9pt}
\caption{Performance of omni models on high-level cinematic failure diagnosis across acting, atmosphere, and cinematography dimensions. Abbreviations: EP = Emotional Performance, MP = Motion Performance, DP = Dialogue Performance, IP = Interaction Performance, MC = Mood Construction, EC = Environmental Coherence, SD = Soundscape Design, IC = Intra-shot Camera, IG = Inter-shot Grammar, and CT = Continuity.}
\vspace{-1ex}
\begin{tabular}{l c cccc ccc ccc c}
\toprule
\multirow{2}{*}{\textbf{Model}} & \multirow{2}{*}{\textbf{Size}}
& \multicolumn{4}{c}{\textbf{Acting Level}}
& \multicolumn{3}{c}{\textbf{Atmosphere Level}}
& \multicolumn{3}{c}{\textbf{Cinematography Level}}
& \multirow{2}{*}{\textbf{Avg.}} \\
\cmidrule(lr){3-6} \cmidrule(lr){7-9} \cmidrule(lr){10-12}
&
& \textbf{EP}
& \textbf{MP}
& \textbf{DP}
& \textbf{IP}
& \textbf{MC}
& \textbf{EC}
& \textbf{SD}
& \textbf{IC}
& \textbf{IG}
& \textbf{CT}
& \\
\midrule

\multicolumn{13}{l}{\textbf{Proprietary Omni Models}} \\
\texttt{Gemini 3.1 Flash Lite} & -- & 48.66 & 27.62 & 49.64 & 38.08 & 32.87 & 40.84 & 66.67 & 33.30 & 49.42 & 56.97 & 44.41 \\
\texttt{Gemini 3.1 Pro} & -- & \underline{53.49} & \textbf{52.35} & \textbf{81.77} & \underline{43.00} & \textbf{71.86} & \textbf{51.57} & \textbf{70.62} & \textbf{49.26} & \textbf{72.68} & \textbf{75.05} & \textbf{62.16} \\
\texttt{Gemini 3 Flash} & -- & 46.11 & 32.21 & \underline{76.45} & \textbf{43.57} & \underline{65.21} & \underline{46.24} & 59.72 & \underline{44.85} & 59.21 & \underline{66.40} & \underline{54.00} \\
\texttt{Gemini 2.5 Flash} & -- & 31.68 & 34.05 & 43.22 & 34.84 & 64.36 & 45.32 & \underline{65.17} & 36.91 & 56.36 & 58.19 & 47.01 \\

\midrule
\multicolumn{13}{l}{\textbf{Open-sourced Omni Models}} \\
\texttt{Qwen 2.5 Omni} & 7B & 35.52 & 23.86 & 40.80 & 33.44 & 34.75 & 31.28 & 45.62 & 32.90 & 51.58 & 48.20 & 37.80 \\
\texttt{MiniCPM-o 2.6} & 7B & 30.05 & 9.01 & 39.73 & 34.04 & 33.51 & 27.64 & 37.93 & 37.26 & \underline{63.01} & 49.15 & 36.13 \\
\texttt{OmniVinci} & 9B & 50.32 & 10.43 & 45.63 & 38.26 & 39.90 & 31.42 & 29.81 & 35.27 & 53.29 & 52.53 & 38.69 \\
\texttt{VideoLLaMA 2} & 7B & \textbf{55.64} & 28.09 & 40.98 & 35.15 & 33.32 & 29.09 & 39.85 & 25.57 & 48.42 & 46.19 & 38.23 \\
\texttt{Ola Omni} & 7B & 31.67 & 14.83 & 39.55 & 34.24 & 36.47 & 31.65 & 56.20 & 33.94 & 59.83 & 43.82 & 37.59 \\
\texttt{Ming Lite Omni 1.5} & 30B & 49.35 & \underline{37.24} & 44.96 & 33.38 & 32.40 & 36.38 & 41.13 & 31.23 & 55.69 & 49.30 & 41.11 \\
\bottomrule
\end{tabular}
\label{tab:omni_results}
\end{table*}

\subsection{QA Subsets Generation}

To transform human-reviewed failure annotations into a benchmark for evaluating omni models, we convert failure cases into structured question-answering (QA) instances that assess failure classification, comparative judgment, and temporal grounding. Given a failure case, its associated failure mode, and supporting evidence, we construct failure-aware diagnostic QA items, where candidate answers are drawn from semantically related failure modes to ensure diagnostic relevance rather than trivial matching. The initial QA construction is assisted by Gemini 3.1 Pro, followed by expert-guided refinement. All QA pairs are further curated through human annotation to improve correctness, discriminability, and difficulty. We adopt a three-way expert validation protocol, in which multiple annotators independently verify and reconcile each QA instance to ensure reliability and consistency. MTAVG-Bench 2.0 includes diverse question formats: single-choice for dominant failure classification, multiple-choice for multi-label or co-occurring failures, pairwise for comparative judgment, and temporal localization for grounding failures in time.

Based on these formats, we organize the benchmark into dedicated subsets for fine-grained failure diagnosis. The final QA set is diverse across categories, sub-dimensions, and question types, covering subtle, confusable, and temporally localized failures, ensuring that strong performance reflects genuine multimodal diagnostic capability rather than overfitting to recurring patterns.

\subsection{Evaluation Protocol}

We evaluate omni large language models on MTAVG-Bench 2.0 under a unified failure-diagnosis setting, measuring their ability to identify high-level cinematic failures from audio-visual inputs. Depending on the subset, model outputs take the form of single-choice, multiple-choice, pairwise judgment, or temporal localization. All outputs are normalized into a structured format via deterministic post-processing; unless otherwise specified, invalid or unparsable responses are treated as incorrect.

For single-choice, pairwise, and multiple-choice questions, we use answer-matching accuracy as the evaluation metric. For temporal localization, we adopt Primary Issue Accuracy (PIA), Temporal Localization Accuracy (TLA), and Rationale Consistency (RC). We report failure-mode QA performance across the three categories of Acting, Atmosphere, and Cinematography, as well as their corresponding sub-dimensions.

\section{Experiments}

\subsection{Experiment Setup}

\textbf{Models.}
We evaluate a diverse set of contemporary omni models that support joint audio--video understanding and multimodal reasoning. Following the benchmark setting in Table~\ref{tab:omni_results}, we include both proprietary and open-source systems to cover different model scales and design paradigms. Specifically, the proprietary group contains \texttt{Gemini 3.1 Flash Lite}\cite{team2023gemini}, \texttt{Gemini 3.1 Pro}, \texttt{Gemini 3 Flash}, and \texttt{Gemini 2.5 Flash}. The open-source group includes \texttt{Qwen 2.5 Omni}\cite{xu2025qwen25omnitechnicalreport} (7B), \texttt{MiniCPM-o 2.6} (7B)\cite{yao2024minicpmvgpt4vlevelmllm}, \texttt{OmniVinci} (9B)\cite{ye2025omnivinci}, \texttt{VideoLLaMA 2} (7B)\cite{cheng2024videollama}, \texttt{Ola Omni} (7B)\cite{liu2025ola}, and \texttt{Ming Lite Omni 1.5} (30B)\cite{ai2025ming}. These models span both lightweight omni systems and larger multimodal architectures, enabling a systematic comparison of high-level cinematic failure diagnosis.

\textbf{Benchmark Setting.} We evaluate all models on MTAVG-Bench 2.0 under a unified failure-diagnosis setting. The benchmark consists of human-reviewed QA instances derived from generated multi-talker audio-video clips, covering three categories—Acting, Atmosphere, and Cinematography—with ten sub-dimensions in total. For each item, the model is given the corresponding video and a question under the storyboard or short-drama setting, targeting failures in a specific sub-dimension. To ensure comparability, we standardize the prompting format and map model outputs into a structured answer space before scoring; unless otherwise specified, invalid or unparsable outputs are counted as incorrect. All models are evaluated in a zero-shot setting with a fixed inference protocol and the original candidate structure of each benchmark item.

\subsection{Evaluation Metrics}

We evaluate models using a hierarchical failure-diagnosis protocol consistent with MTAVG-Bench 2.0 and Table~\ref{tab:omni_results}. Each question is assigned to one of ten fine-grained sub-dimensions under three categories: Acting, Atmosphere, and Cinematography. Accuracy is computed per question based on task format, using exact match for single-choice and pairwise questions, and answer consistency for multiple-choice questions. Scores are averaged within each sub-dimension, and the final Avg. is the weighted mean over all ten sub-dimensions, preventing any category from dominating due to differences in question count. This protocol emphasizes high-level cinematic failure diagnosis from audio-video inputs and supports both fine-grained and holistic evaluation across models.

\subsection{Main Result}

Table~\ref{tab:omni_results} shows that proprietary omni models consistently outperform open-source models on MTAVG-Bench 2.0. Among all evaluated systems, \texttt{Gemini 3.1 Pro} achieves the best overall performance with an average score of \textbf{62.16}, substantially higher than the strongest open-source model, \texttt{Ming Lite Omni 1.5} (41.11). Its advantage is particularly pronounced in the \textbf{Atmosphere} and \textbf{Cinematography} dimensions, where it obtains the best results on Mood Construction (71.86), Environmental Coherence (51.57), Soundscape Design (70.62), Intra-shot Camera (49.26), Inter-shot Grammar (72.68), and Continuity (75.05). In the \textbf{Acting} category, performance is more mixed: \texttt{Gemini 3.1 Pro} leads on Motion Performance (52.35) and Dialogue Performance (81.77), while \texttt{Gemini 3 Flash} performs best on Interaction Performance (43.57), and \texttt{VideoLLaMA 2} achieves the highest Emotional Performance score overall (55.64). These results suggest that MTAVG-Bench 2.0 clearly separates model capability on high-level cinematic failure diagnosis: current proprietary models show substantially stronger scene-level reasoning ability, while open-source models exhibit only isolated strengths on a few sub-dimensions.

\section{Analysis}

\subsection{Model-wise Failure Rate Analysis}
\begin{figure}
    \centering
    \includegraphics[width=1\linewidth]{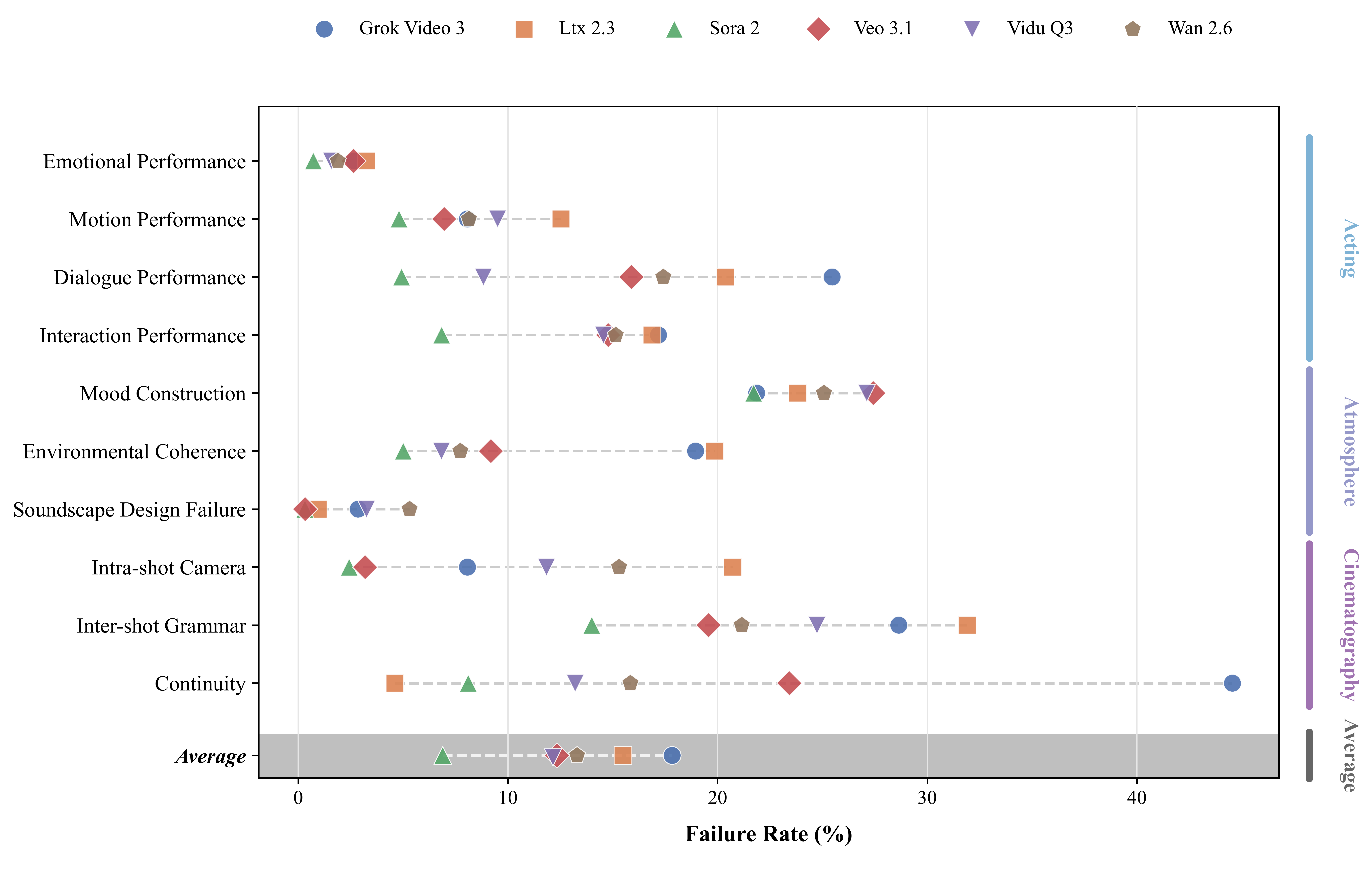}
    \caption{\textbf{Failure rates across fine-grained failure modes for different video generation models.}
    The figure shows the percentage of generated videos exhibiting each failure type across the ten sub-dimensions of \textbf{acting}, \textbf{atmosphere}, and \textbf{cinematography}. Each marker represents a model, and the shaded bottom row shows the average failure rate across all dimensions. Lower values indicate better performance.}
    \label{fig:failure_generation}
\end{figure}

Figure~\ref{fig:failure_generation} compares failure rates across the ten fine-grained sub-dimensions for different video generation models. Two observations are particularly noteworthy. First, failure patterns are highly uneven across dimensions. Compared with relatively localized failures, such as emotional performance or soundscape design, dimensions requiring richer scene-level coordination---including dialogue performance, interaction performance, mood construction, inter-shot grammar, and continuity---tend to exhibit consistently higher failure rates across multiple models. This suggests that high-level scene generation remains especially challenging when successful performance depends on coordinated acting, atmosphere construction, and cross-shot cinematic organization.

Second, model robustness varies substantially across sub-dimensions. No single system is uniformly reliable across all failure types: models that perform relatively well on acting-related dimensions may still struggle on cinematography-related ones, and vice versa. In particular, \texttt{Grok Video 3} and \texttt{LTX 2.3} exhibit relatively high failure rates across multiple sub-dimensions, indicating persistent weaknesses in scene-level cinematic generation. The shaded average row further shows that even the strongest video generators still suffer from non-trivial failure rates overall. These results highlight the need for fine-grained diagnostic evaluation, since aggregate quality judgments alone would obscure the heterogeneous weaknesses of current systems.

\subsection{Holistic Automatic Quality Assessment of Video Sources}
Table~\ref{tab:video_evaluation} reports a holistic quality assessment of different video sources using deep learning--based automatic metrics. The results reveal complementary strengths across models rather than a single dominant system. For example, \texttt{Grok Video 3} achieves the best performance on \textit{Audio Aesthetic}, \textit{T-A Align}, and \textit{T-V Align}, whereas \texttt{Veo 3.1} performs best on \textit{Lip Sync} and \textit{A-V Align}. Meanwhile, \texttt{LTX 2.3} obtains the lowest \textit{Desync}, suggesting stronger temporal synchronization under this metric.

However, these automatic metrics do not fully capture the high-level failures revealed in Figure~\ref{fig:failure_generation}. A representative example is \texttt{Grok Video 3}: although it performs well on several audio-related objective metrics, it still exhibits relatively high failure rates in this diagnostic benchmark. A closer inspection suggests that this discrepancy is partly due to its tendency to generate acoustically clean and perceptually pleasing speech signals, which can improve low-level audio and aesthetic scores, while still lacking sufficient environmental grounding and contextual integration. As a result, the generated audio may sound clean in isolation but remain weak in environmental coherence and scene-level soundscape consistency. More broadly, models with strong low-level fidelity or alignment scores can still exhibit substantial weaknesses in acting, atmosphere, or cinematography. This discrepancy underscores the central motivation of MTAVG-Bench 2.0: conventional automatic metrics are useful for measuring perceptual fidelity and alignment quality, but they remain insufficient for diagnosing scene-level cinematic failures in a fine-grained and interpretable manner.

\begin{table}[t]
\centering
\large
\setlength{\tabcolsep}{6pt}
\renewcommand{\arraystretch}{1.1}
\caption{Comparison of different video sources under a holistic quality assessment protocol using deep learning--based automatic evaluation metrics. Best results are in \textbf{bold} and second-best results are \underline{underlined}.}
\label{tab:video_evaluation}
\resizebox{\columnwidth}{!}{%
\begin{tabular}{lcccccc}
\toprule
\textbf{Video Source} &
\makecell{\textbf{Audio}\textbf{Aesthetic}$\uparrow$}&
\makecell{\textbf{Lip}\textbf{Sync}$\uparrow$}&
\makecell{\textbf{A-V}\textbf{Align}$\uparrow$}&
\makecell{\textbf{Desync}$\downarrow$} &
\makecell{\textbf{T-A}\textbf{ Align}$\uparrow$}&
\makecell{\textbf{T-V}\textbf{Align}$\uparrow$}\\
\midrule
Grok Video 3 & \textbf{4.283} & 0.682 & 0.160 & 0.614 & \textbf{0.349} & \textbf{0.222} \\
Ltx 2.3      & 3.942 & 0.454 & 0.117 & \textbf{0.377} & 0.170 & 0.191 \\
Sora 2       & 2.614 & 0.438 & 0.187 & 0.593 & 0.158 & \underline{0.215} \\
Veo 3.1      & 3.626 & \textbf{1.116} & \textbf{0.235} & 0.600 & 0.237 & 0.211 \\
Vidu Q3      & 4.108 & \underline{0.952} & \underline{0.194} & 0.635 & 0.204 & 0.204 \\
Wan 2.6      & \underline{4.241} & 0.458 & 0.128 & \underline{0.545} & \underline{0.260} & 0.211 \\
\bottomrule
\end{tabular}%
}
\end{table}

\begin{table}[h]
\centering
\small
    \caption{Performance on Timestamp Failure Mode Localization. \textbf{PIA}, \textbf{TLA}, and \textbf{RC} denote \textbf{Primary Issue Accuracy}, \textbf{Temporal Localization Accuracy}, and \textbf{Rationale Consistency}, respectively.}
     \label{tab:timestamp_failure_localization}
     \renewcommand{\arraystretch}{1}
     \begin{tabular}{lccc}
         \toprule
         \textbf{Model} & \textbf{PIA} & \textbf{TLA} & \textbf{RC} \\
         \midrule
         Gemini 3.1 Pro & 60.6\% & 60.9\% & 83.8\% \\
         \bottomrule
     \end{tabular}
 \end{table}

\subsection{Timestamp Failure Mode Localization}
Table~\ref{tab:timestamp_failure_localization} reports performance on timestamp failure mode localization. Compared with standard failure recognition, this setting is more demanding as the evaluator must identify both what failure occurs and where it happens in time. We evaluate Gemini 3.1 Pro to probe the upper bound of current temporal capabilities. Notably, while the model achieves a high Rationale Consistency (RC) of $83.8\%$, its Primary Issue Accuracy (PIA) and Temporal Localization Accuracy (TLA) drop significantly to $60.6\%$ and $60.9\%$, respectively. This stark contrast reveals a critical limitation of current omni models: although they can generate plausible textual rationales for cinematic failures, they struggle to precisely anchor these visual or auditory issues to specific timestamps. This indicates that fine-grained temporal grounding remains a major bottleneck for scene-level diagnosis.

\subsection{Ablation Study on Input Modalities}
Table~\ref{tab:scene-level-ablation} presents an input ablation study on Gemini 3 Flash to investigate the reliance of cinematic diagnosis on multimodal reasoning. The full audio-visual input yields the best average performance ($54.00\%$). Removing the visual modality causes a drastic drop to $33.10\%$, barely exceeding the Text-only baseline ($32.82\%$). This proves that evaluating Acting and Cinematography heavily relies on visual cues like expressions and framing. Conversely, removing audio reduces the average score to $40.64\%$. Notably, this absence not only degrades the Atmosphere dimension but also impairs Acting and Cinematography, emphasizing the necessity of audio-visual synchrony and rhythm. Overall, these findings confirm that MTAVG-Bench 2.0 strictly requires joint multimodal understanding, effectively preventing models from exploiting unimodal shortcuts.

\subsection{Case Study of Diagnostic QA}
\begin{figure}
    \centering
    \includegraphics[width=1\linewidth]{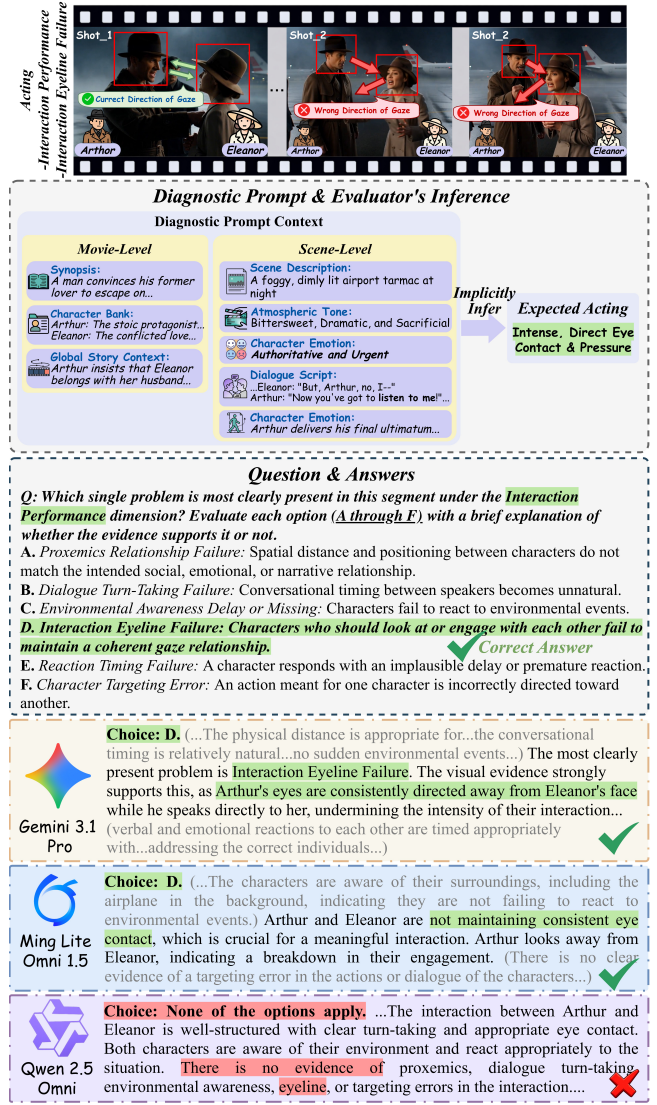}
    \caption{\textbf{Case study of diagnostic QA for a failure case under Interaction Performance.}}
    \label{fig:case}
    \vspace{-2ex}
\end{figure}

Figure~\ref{fig:case} presents a representative diagnostic QA example under the \textit{Interaction Performance} dimension. In this scene, the prompt implies a tense, confrontational exchange in which direct eye contact is expected. However, the generated segment exhibits inconsistent gaze behavior, making \textit{Interaction Eyeline Failure} the correct diagnosis. This example shows that successful evaluation requires more than simply recognizing that a clip is problematic. The evaluator must infer the intended acting from context, distinguish among closely related failure types, and ground the answer in concrete visual evidence. As shown in the lower part of the figure, stronger evaluators correctly identify the eyeline inconsistency and justify their answers with scene-specific evidence, whereas weaker ones miss the key cue and produce an incorrect diagnosis.

\begin{table}[t]
    \centering
    \small 
    \caption{Performance of \textbf{Gemini 3 Flash} on scene-level failure mode identification under different input settings. We compare the full multimodal setting with three ablations: w/o vision, w/o audio, and text-only.}
    \label{tab:scene-level-ablation}
    
    \renewcommand{\arraystretch}{0.9} 
    
    \setlength{\tabcolsep}{6pt} 
    
    \begin{tabular}{lcccc} 
        \toprule
        \textbf{Input Setting} & \textbf{Acting} & \textbf{Atmosphere} & \textbf{Cinema.} & \textbf{Avg.} \\
        \midrule
        Full Input & 49.58\% & 57.06\% & 56.82\% & 54.00\% \\
        w/o Audio  & 36.40\% & 41.20\% & 45.73\% & 40.64\% \\
        w/o Vision & 36.67\% & 33.70\% & 27.73\% & 33.10\% \\
        Text-only  & 34.88\% & 33.10\% & 29.80\% & 32.82\% \\
        \bottomrule
    \end{tabular}
\end{table}

\section{Conclusion}

We introduce \textbf{MTAVG-Bench 2.0}, a benchmark for diagnosing high-level failures in multi-talker audio-video generation. Unlike conventional evaluation that primarily focuses on low-level fidelity, synchronization, and local interaction quality, MTAVG-Bench 2.0 formulates scene-level cinematic evaluation as a structured diagnosis problem. To this end, we establish a failure taxonomy spanning acting\textbf{,} atmosphere\textbf{,} and cinematography, and construct over 11,000 question-answering instances, along with dedicated subsets for scene-level evaluation and temporal localization. Experimental results show that leading commercial omni models achieve the strongest overall performance, yet still exhibit clear limitations on challenging cases that require fine-grained reasoning about expressive acting, atmosphere construction, cinematic grammar, and temporally grounded evidence. Further analysis reveals that strong performance on conventional automatic quality metrics does not necessarily translate into reliable scene-level failure diagnosis. These findings indicate that evaluating cinematic multi-talker generation requires a richer framework beyond synchronization, alignment, and perceptual quality. Overall, MTAVG-Bench 2.0 provides a systematic benchmark for studying the capability boundary of current omni models in cinematic failure diagnosis, and lays the foundation for interpretable, film-level evaluation of audio-video generation.

\clearpage
\bibliographystyle{ACM-Reference-Format}
\bibliography{ref}

\newcommand{\appendixasinput}{}
\ifdefined\appendixasinput
\else
  \documentclass[sigconf,screen]{acmart}

  \settopmatter{
    printacmref=false,
    printccs=false,
    printfolios=false
  }
  \renewcommand\footnotetextcopyrightpermission[1]{}

  \usepackage{fancyhdr}
  \usepackage[most]{tcolorbox}
  \usepackage{listings}

  \newtcblisting{systemprompt}[1][]{%
    enhanced,
    breakable,
    listing only,
    listing options={
      basicstyle=\ttfamily\scriptsize,
      breaklines=true,
      breakatwhitespace=false,
      columns=fullflexible,
      keepspaces=true,
      showstringspaces=false,
    },
    colback=blue!5,
    colframe=blue!55!black,
    colbacktitle=blue!55!black,
    coltitle=white,
    title=System Instruction,
    boxrule=0.5pt,
    arc=2mm,
    left=8pt,
    right=8pt,
    top=8pt,
    bottom=8pt,
    fonttitle=\bfseries,
    #1
  }

  \AtBeginDocument{%
    \providecommand\BibTeX{{Bib\TeX}}%
  }

  \begin{document}

  \makeatletter
  \fancypagestyle{plain}{%
    \fancyhf{}%
    \renewcommand{\headrulewidth}{0pt}%
    \renewcommand{\footrulewidth}{0pt}%
  }
  \fancypagestyle{standardpagestyle}{%
    \fancyhf{}%
    \renewcommand{\headrulewidth}{0pt}%
    \renewcommand{\footrulewidth}{0pt}%
  }
  \makeatother

  \pagestyle{plain}
  \thispagestyle{plain}
  \markboth{}{}
  \twocolumn
\fi


\clearpage
\appendix

\pagestyle{plain}
\thispagestyle{plain}
\markboth{}{}

\section*{Appendix}

\section{Prompt Design for Benchmark Construction and Evaluation}
We designed a set of structured system prompts to support scene decomposition, segment-level video generation, and later model evaluation. These prompts define the required roles, field structures, and output constraints for different stages of the benchmark pipeline.

\subsection{Shot Analysis Prompts}
To convert uploaded movie clips into de-identified structured plans for downstream generation, we use a dedicated shot-analysis system prompt. This prompt asks the model to produce a copyright-safe JSON plan with clip-level summary fields, inference rules, and 8-to-10-second segments. Because the full template is long, we present it across two consecutive pages in Figures~\ref{fig:shot_analysis_prompt_box_1} and~\ref{fig:shot_analysis_prompt_box_2}.

\subsection{Video Generation Prompts}
To synthesize segment-level multi-talker videos from structured script metadata, we use a dedicated video-generation system prompt. This prompt consumes the de-identified fields produced by the shot-analysis stage and enforces continuity, dialogue fidelity, and multi-shot cinematic structure, as shown in Figure~\ref{fig:video_generation_prompt_box}.

\subsection{Evaluation Prompts}
For rationale-consistency evaluation, we use a shared judge prompt across GPT-5.4. The prompt asks each judge model to assess whether the predicted rationale logically supports the final answer choice and to return a structured Likert-scale score together with a short analysis, as shown in Figure~\ref{fig:rc_judge_prompt_box}.

\subsection{Prompt Variants and Output Formats}
This section documents prompt revisions, structured JSON output requirements, and the main output-format constraints shared across the evaluation pipeline. The full prompt templates are reproduced on the following pages.

\section{Benchmark Construction Details}
\subsection{Data Source and Sample Collection}
We first use Gemini 3.1 Pro to perform scene-level analysis on source videos and produce draft descriptions. Human annotators then refine these drafts so that they better match the original film content. Before QA construction, each retained case is associated with the video segment, failure evidence, script or dialogue information, temporal cues, and a fixed taxonomy label.

\begin{figure}[t]
\centering
\includegraphics[width=\columnwidth]{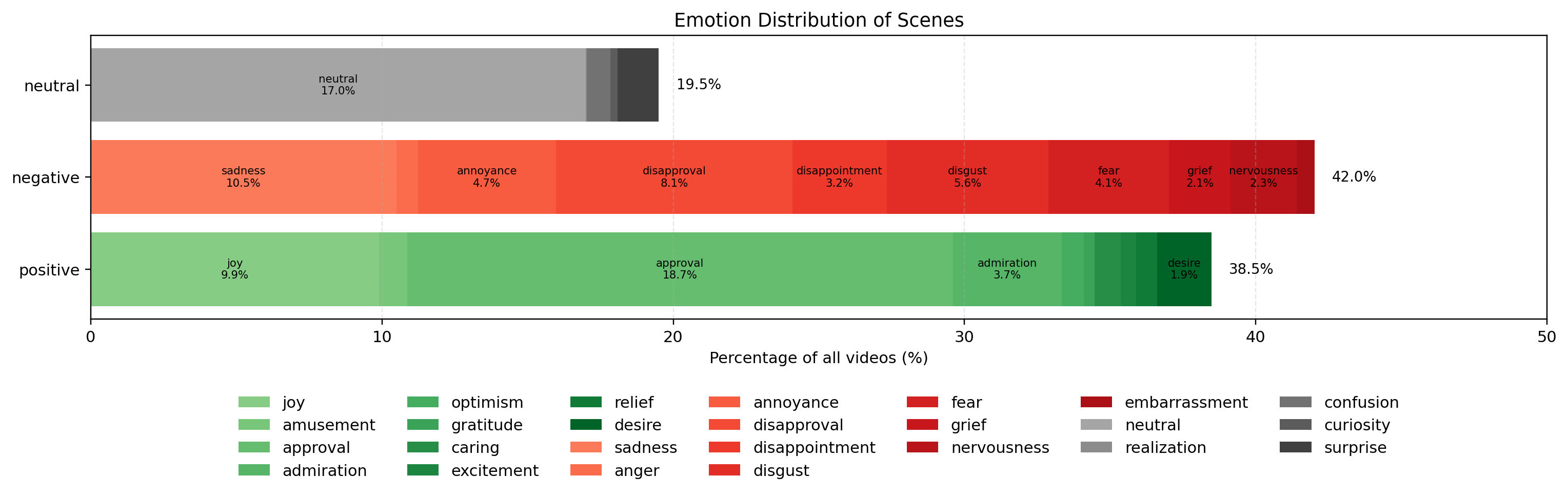}
\caption{Stacked sentiment composition across positive, negative, and neutral rows.}
\label{fig:sentiment_distribution}
\end{figure}

We further analyze the coarse emotion composition of the 2,466 scene-level samples drawn from 20 source movie clips. As shown in Figure~\ref{fig:sentiment_distribution}, the overall distribution is reasonably balanced across the three sentiment rows, with negative scenes accounting for 42.0\% of all samples, positive scenes accounting for 38.5\%, and neutral scenes accounting for 19.5\%. This suggests that the dataset is not dominated by a single emotional polarity and instead preserves broad coverage of positive, negative, and neutral dialogue situations.

Within the positive row, approval is the largest component at 18.7\%, followed by joy at 9.9\% and admiration at 3.7\%. Within the negative row, sadness (10.5\%) and disapproval (8.1\%) are the most prevalent categories, followed by disgust (5.6\%), annoyance (4.7\%), and fear (4.1\%). The neutral row is centered primarily on neutral affect itself (17.0\%), with only a small residual contribution from adjacent low-intensity states. Overall, this composition indicates that the benchmark covers a broad range of interpersonal emotional settings while still emphasizing tension, conflict, and evaluative exchange, which are common in dialogue-centric cinematic scenes.

\subsection{Dialogue and Script Construction Pipeline}
The script-construction pipeline converts coarse scene analysis into structured generation inputs. During human refinement, annotators complete missing dialogue, reorder shots or events when necessary, and align character actions with the original movie. The resulting representation is a structured script description rather than a plain transcript. It may include premise, core setting, overall tone, character registry, dialogue, emotion, and other scene-level cues used for generation.

\subsection{Filtering and Refinement Procedure}
We exclude cases that do not have an explicit failure-mode annotation before QA construction. For the remaining cases, refinement focuses on keeping the script and evidence aligned with the source scene so that later questions are grounded in a concrete failure label. This step reduces loosely specified errors and preserves the diagnostic role of each item.

\subsection{Question Template Construction}
Questions are constructed with a rule-based pipeline rather than written manually item by item. We maintain more than twenty templates covering single-choice, multiple-choice, pairwise, and temporal-localization formats. Templates are matched with failure modes without additional hand-crafted constraints, and candidate options are generated from the corresponding failure-mode candidate set. This design supports failure-aware diagnosis while preserving task-format diversity.

\section{Annotation Quality Control}
\subsection{Annotation and Verification Rules}
We use an expert pool of 22 human verifiers for QA refinement and validation. For each item, two experts are first sampled from this pool for independent review. The core judgment is whether the video actually exhibits the target failure under the fixed taxonomy. Initial agreement is therefore determined by whether the two reviewers assign the same failure mode. Under this protocol, the initial two-expert review reaches an agreement rate of 84.1\%, corresponding to a Cohen's kappa of 0.78.

\subsection{Failure Case Filtering}
The main filtering rule at this stage is simple: cases without an explicit failure-mode annotation are removed before benchmark construction. For retained items, revision is still allowed when the generated question-answer pair does not faithfully reflect the labeled failure. This keeps the benchmark focused on cases with clear diagnostic targets.

\clearpage

\begin{figure*}[p]
\centering
\begin{systemprompt}[title=Shot Analysis Prompt (Part I)]
You are an expert film analyst and content transformer. Your task is to analyze the uploaded film clip and convert it into a copyright-safe, neutral, structured JSON plan for chained video generation.

Operational Step (Internal): First, internally identify the original film, the specific scene, and the characters. Use this identified context to fill the premise and segments fields, but strictly ensure that the final output contains no copyrighted names, titles, or IP-specific references.

Important Goals:
- The output will be used for multi-segment video generation, one segment at a time.
- Each generated segment will later be turned into a prompt automatically from the JSON keys, so the JSON must be rich, precise, and internally consistent.
- Output JSON only. Do not output markdown, explanations, or extra text.
- Do not identify or mention the original film title, franchise, copyrighted character names, copyrighted place names, or any IP-specific proper nouns from the source clip.
- Do not use placeholder labels like A, B, C. Instead, assign invented names that fit the apparent gender and age of each character, and keep those names consistent across the whole JSON.
- Rewrite any profanity, trademarked references, or copyright-sensitive phrases into neutral alternatives while preserving scene meaning and approximate dramatic intent.

Segmentation Rules:
- Split the clip into sequential 8.0-second generation segments.
- For each segment, set: source_start_sec, source_end_sec, target_duration_sec, reference_frame_timestamp_sec.
- Use 8.0 as target_duration_sec for each segment.
- Divide spoken dialogue across segments so that each segment contains only the dialogue that belongs in that segment.
- If the source clip does not divide perfectly, still organize it into usable 8-second generation units without changing the core scene logic.

Critical Dialogue Rule:
- The dialogue field is the canonical spoken content for that segment.
- When characters speak, the spoken dialogue must exactly match the lines provided in the dialogue field for that segment.
- Do not add, omit, paraphrase, reorder, or summarize spoken words inside dialogue.

Style and Inference Rules:
- Do not explicitly direct lighting, color palette, background music, facial expression, body language, or camera choreography unless plot-critical.
- Encode those as model inference rules so the downstream video model can infer them.
- The scene should be designed for a suitable multi-shot result, not a single static shot.
- Do not show any subtitles in any frame.

Output Requirements:
- Return a single valid JSON object with exactly these top-level keys:
  - clip_summary
  - model_inference_rules
  - segments

Schema (Fill with content inferred and de-identified from the clip):
JSON
{
  "clip_summary": {
    "premise": "Based on the identified scene, provide one concise sentence describing the core dramatic situation using neutral language.",
    "core_setting": "",
    "overall_tone": "",
    "character_registry": {
      "InventedName1": {
        "role": "",
        "persistent_identity": "",
        "relationship_summary": ""
      }
    }
  },
  "model_inference_rules": {
    "visual_atmosphere_rule": "Infer lighting, color, and overall visual atmosphere from the emotional tone and story situation. Do not require explicit visual styling instructions unless plot-critical.",
    "music_rule": "Infer whether background music is needed, and its emotional character, from the dramatic tone of the segment.",
    "performance_rule": "Infer facial expression, body language, and paralanguage from the emotional tone, relationship dynamic, and spoken dialogue.",
    "dialogue_rule": "When characters speak, the spoken dialogue must exactly match the lines provided in the dialogue field for this segment. Do not add, omit, paraphrase, or reorder any spoken words.",
    "restraint_rule": "Avoid exaggerated performance or visual stylization unless strongly justified by the scene.",
    "continuity_rule": "Maintain continuity of character identity, environment, and carry-over dramatic state from the previous segment.",
    "camera_rule": "This segment must be realized as a multi-shot scene with clear shot progression, rather than a single continuous static view (A cinematic structure in which a single scene or narrative moment is constructed from multiple shots that appear sequentially over time, with each shot presenting a different camera angle, framing, or perspective). Freely infer camera movement, framing, shot transitions, and visual pacing from the script, dialogue, emotional tone, relationship dynamics, and dramatic context."
  },


\end{systemprompt}
\caption{System prompt for scene-level shot analysis and de-identified JSON planning (Part I).}
\label{fig:shot_analysis_prompt_box_1}
\end{figure*}
\clearpage

\begin{figure*}[p]
\centering
\begin{systemprompt}[title=Shot Analysis Prompt (Part II)]

  "segments": [
    {
      "segment_id": "seg_001",
      "source_start_sec": 0.0,
      "source_end_sec": 8.0,
      "target_duration_sec": 8.0,
      "reference_frame_timestamp_sec": 0.0,
      "story_context": "",
      "environment": "",
      "characters_present": [],
      "dialogue": [
        {
          "speaker": "",
          "line": "",
          "action": ""
        }
      ],
      "emotional_tone": "",
      "relationship_dynamic": "",
      "key_event": "",
      "continuity_from_previous": ""
    }
  ]
}

Field guidance:
- premise: Based on the identified scene, provide one concise sentence describing the core dramatic situation using neutral language.
- core_setting: stable physical setting for the clip.
- overall_tone: stable emotional/dramatic tone for the clip.
- role: dramatic function in the scene, written neutrally.
- persistent_identity: stable, visible appearance cues only; no copyrighted identifiers.
- relationship_summary: concise relationship description relevant for performance and continuity.
- story_context: what is happening in this segment, in plain neutral language.
- environment: only the plot-relevant environment for this segment.
- characters_present: list of invented character names present in the segment.
- dialogue: exact rewritten spoken lines for this segment only.
- action: character behavior motivated by narrative progression rather than emotional expression.
- emotional_tone: short phrase describing the dominant emotional state.
- relationship_dynamic: short phrase describing the interpersonal tension or bond in this segment.
- key_event: the segment's main dramatic beat.
- continuity_from_previous: what emotional or narrative state carries over from the previous segment.

Additional constraints:
- Do not include video_generation_prompt.
- Do not mention the original movie, franchise, or actor names.
- Do not mention "the original film", "the source movie", or similar meta language inside the JSON.
- Do not use bullet formatting inside string values.
- Keep the JSON compact but sufficiently descriptive for downstream prompt construction.
- Use stable invented names that fit the apparent gender of each character.
- If a segment has no spoken dialogue, set "dialogue": [].
- Output valid JSON only.
\end{systemprompt}
\caption{System prompt for scene-level shot analysis and de-identified JSON planning (Part II).}
\label{fig:shot_analysis_prompt_box_2}
\end{figure*}
\clearpage

\begin{figure*}[p]
\centering
\begin{systemprompt}[title=Video Generation Prompt]

You are a cinematic multi-talker segment expert. Generate one multi-shot video segment.

Global clip context:
- Premise: {premise}
- Core setting: {core_setting}
- Overall tone: {overall_tone}

Character registry:
- {character_registry}

Current segment:
- Source start time: {source_start_sec}
- Source end time: {source_end_sec}
- Target duration: {target_duration_sec}
- Reference frame timestamp: {reference_frame_timestamp_sec}
- Story context: {story_context}
- Environment: {environment}
- Characters present: {characters_present}
- Emotional tone: {emotional_tone}
- Relationship dynamic: {relationship_dynamic}
- Key event: {key_event}
- Continuity from previous: {continuity_from_previous}

Dialogue:
- Use exactly the dialogue entries provided for this segment.
- If the segment has no spoken dialogue, use no spoken dialogue.
- Dialogue entries:
{dialogue}

Generation objective:
- Realize this scene from the provided script metadata rather than copying any known film-specific phrasing or IP-specific world details.
- Preserve character identity, relationship tension, and story continuity across segments.
- Keep the scene grounded, emotionally restrained, and driven by the dialogue and dramatic situation.

Reference handling:
- If a reference image is provided, use it only to preserve character, costume, environment, and dramatic continuity from the immediately preceding segment.
- The reference image should correspond to the current segment's reference_frame_timestamp_sec.
- Do not copy the exact composition of the reference image unless it is necessary for continuity.
- If no reference image is provided, rely on the global clip context, character registry, and current segment metadata.

Model inference rules:
- Infer lighting, color, and overall visual atmosphere from the emotional tone and story situation. Do not require explicit visual styling instructions unless plot-critical.
- Infer whether background music is needed, and its emotional character, from the dramatic tone of the segment.
- Infer facial expression, body language, and paralanguage from the emotional tone, relationship dynamic, and spoken dialogue.
- When characters speak, the spoken dialogue must exactly match the lines provided in the dialogue field for this segment. Do not add, omit, paraphrase, reorder, or summarize any spoken words.
- If the segment contains no dialogue, do not invent speech.
- Avoid exaggerated performance or visual stylization unless strongly justified by the scene.
- Maintain continuity of character identity, environment, and carry-over dramatic state from the previous segment.
- Realize the segment as a cinematic sequence composed of multiple shots connected by editing cuts.
- Use sequential shot changes over time, such as wide shot, medium shot, close-up, or over-the-shoulder framing, when appropriate to the scene.
- The video must always remain a single full-frame image.
- Do not use split screen, multi-panel layouts, grids, picture-in-picture, collage, or multiple simultaneous views within one frame.
\end{systemprompt}
\caption{System prompt for segment-level video generation.}
\label{fig:video_generation_prompt_box}
\end{figure*}
\clearpage

\begin{figure*}[p]
\centering
\begin{systemprompt}[title=RC Judge Prompt]
You are a rigorous reasoning evaluation expert. Evaluate whether the model's rationale logically supports its final answer choice.

[Question]
{question}

[Options]
{options}

[Model's Predicted Option]
{predicted_option_id}: {predicted_option_text}

[Model's Rationale]
{rationale}

Evaluation Criteria (Likert Scale 1-5):
- 1: Rationale completely contradicts or is irrelevant to the chosen option
- 2: Rationale shows some relevance but contains significant logical gaps
- 3: Rationale moderately supports the chosen option with minor issues
- 4: Rationale well supports the chosen option with clear reasoning
- 5: Rationale fully and logically supports the chosen option

Output Format (JSON):
{{"score": <integer 1-5>, "rationale": "<detailed analysis>"}}
\end{systemprompt}
\caption{Shared judge prompt used for rationale-consistency evaluation.}
\label{fig:rc_judge_prompt_box}
\end{figure*}
\clearpage

\subsection{Question Template Construction}
Questions are constructed with a rule-based pipeline rather than written manually item by item. We maintain more than twenty templates covering single-choice, multiple-choice, pairwise, and temporal-localization formats. Templates are matched with failure modes without additional hand-crafted constraints, and candidate options are generated from the corresponding failure-mode candidate set. This design supports failure-aware diagnosis while preserving task-format diversity.

\section{Annotation Quality Control}
\subsection{Annotation and Verification Rules}
We use an expert pool of 22 human verifiers for QA refinement and validation. For each item, two experts are first sampled from this pool for independent review. The core judgment is whether the video actually exhibits the target failure under the fixed taxonomy. Initial agreement is therefore determined by whether the two reviewers assign the same failure mode. Under this protocol, the initial two-expert review reaches an agreement rate of 84.1\%, corresponding to a Cohen's kappa of 0.78.

\subsection{Failure Case Filtering}
The main filtering rule at this stage is simple: cases without an explicit failure-mode annotation are removed before benchmark construction. For retained items, revision is still allowed when the generated question-answer pair does not faithfully reflect the labeled failure. This keeps the benchmark focused on cases with clear diagnostic targets.

\subsection{Conflict Resolution}
If the first two experts agree on the target failure mode, the item is accepted directly. If disagreement arises, a third expert from the same pool is introduced for discussion-based adjudication. In practice, this stage is mainly used to determine which failure mode best describes the video. This third-expert intervention is required for 15.9\% of items. The taxonomy-to-label mapping remains fixed throughout the process: each of the 45 fine-grained failure modes belongs to one of 10 sub-dimensions, which are further grouped into 3 top-level categories.

\section{Evaluation Protocol Details}
\subsection{Model Inference Settings}
Closed-source models are evaluated through their official APIs, while open-source models are run locally. Unless required by backend constraints, we keep decoding settings conservative and broadly consistent across models to reduce unnecessary variance during evaluation. Model-specific adjustments are introduced only when needed to satisfy interface, context-length, or serving requirements.

\subsection{Sampling Strategy}
Each evaluation item is defined on a scene-level short video rather than an arbitrary crop. For temporal localization, timestamps are generated in a rule-based manner from these scene-level annotations. This keeps the temporal target aligned with the same segmentation used during failure annotation.

\subsection{Judgment and Parsing Rules}
We unify different task formats under a normalized per-question scoring scheme. For each question $i$, let $G_i$ denote the ground-truth answer set and $P_i$ denote the model prediction. This design prioritizes coverage of ground-truth failure evidence over precision of option selection. The per-question score $s_i \in [0,1]$ is defined separately for the three QA formats used in the benchmark:
\begin{equation}
s_i = \mathbb{I}[P_i = G_i],
\end{equation}
for single-choice questions.
\begin{equation}
s_i = \frac{|P_i \cap G_i|}{|G_i|},
\end{equation}
for multiple-choice questions.
\begin{equation}
s_i = \mathbb{I}[P_i = G_i],
\end{equation}
for pairwise comparison questions. Thus, single-choice and pairwise questions use exact-match scoring, while multiple-choice questions are evaluated by the fraction of ground-truth options covered by the prediction. For each sub-dimension $d$, the score is computed as
\begin{equation}
S_d = \frac{1}{N_d} \sum_{i \in \mathcal{Q}_d} s_i,
\end{equation}
where $\mathcal{Q}_d$ is the set of questions in sub-dimension $d$ and $N_d = |\mathcal{Q}_d|$. The overall score is then obtained as a weighted average over all sub-dimensions:
\begin{equation}
\mathrm{Avg.} = \sum_{d=1}^{D} \frac{N_d}{\sum_{k=1}^{D} N_k} S_d.
\end{equation}
Unless otherwise specified, invalid or unparsable model outputs are treated as incorrect during evaluation.

For timestamp localization, we report Primary Issue Accuracy (PIA), Temporal Localization Accuracy (TLA), and Rationale Consistency (RC). PIA measures whether the predicted primary failure label matches the annotated primary failure label:
\begin{equation}
\mathrm{PIA} = \frac{1}{N} \sum_{i=1}^{N} \mathbb{I}[\hat{y}_i = y_i].
\end{equation}
TLA measures whether the predicted timestamp options cover the annotated failure region, using the same coverage-based scoring rule as in multiple-choice QA:
\begin{equation}
\mathrm{TLA} = \frac{1}{N} \sum_{i=1}^{N} \frac{|\hat{T}_i \cap T_i|}{|T_i|}.
\end{equation}
RC measures how well the model-generated rationale aligns with the annotated failure evidence, temporal localization, and final failure diagnosis. We use GPT-5.4 as judge under the same standardized evaluation prompt. Each judge assigns a 1--5 Likert score based on the alignment of the rationale with the failure evidence, the temporal localization, and the final failure diagnosis. We average the three scores and convert the result into a percentage scale:
\begin{equation}
\mathrm{RC} = \frac{100}{5N} \sum_{i=1}^{N} r_i,
\end{equation}
where $r_i \in [0,5]$ denotes the average score assigned by the three judge models to instance $i$.

\subsection{Failure Rate Definition}
In addition to the task-specific metrics above, we analyze generation-side failure incidence at the clip-segment level. For a video generation model $m$, let $V_m$ denote the full set of generated videos and let $|V_m|$ be the total number of generated videos. For a failure dimension $d$, let $C_{m,d}$ denote the number of annotated failure clip segments assigned to that dimension in videos generated by model $m$. We define the corresponding failure rate as
\begin{equation}
\mathrm{FR}_{m,d}
= \frac{C_{m,d}}{|V_m|}.
\end{equation}
When multiple failures occur within the same video, they are accumulated separately in their corresponding dimensions. Therefore, this statistic reflects clip-segment-level failure incidence normalized by the total number of generated videos, rather than a binary video-level failure indicator.

\section{Additional Quantitative Results}

\subsection{Per-Failure-Mode Results}
\begin{figure*}[p]
\centering
\includegraphics[height=0.9\textheight,keepaspectratio]{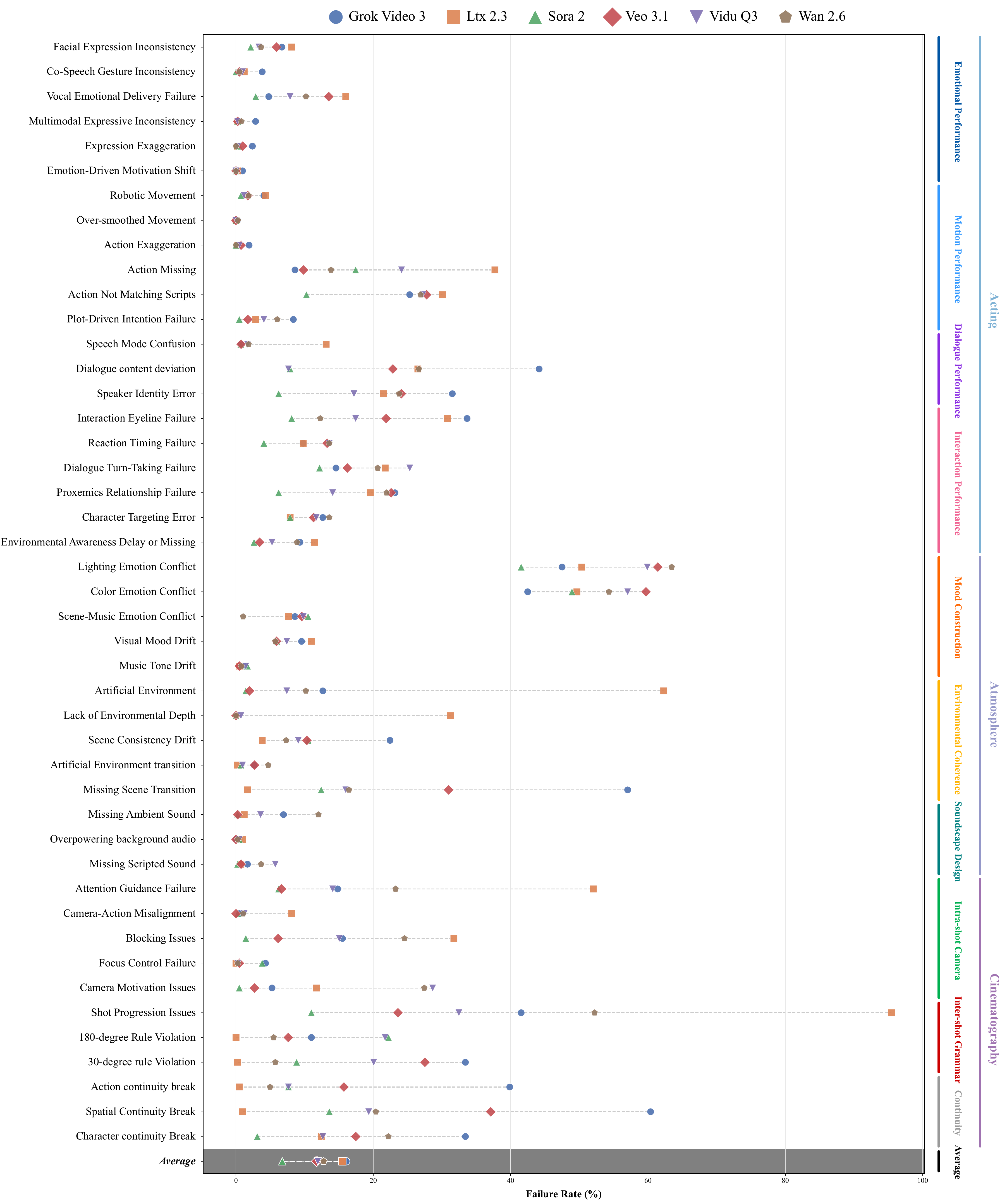}
\caption{Failure Rate on each failure mode}
\label{fig:failure_rate_per_mode}
\end{figure*}

Figure~\ref{fig:failure_rate_per_mode} reveals a highly non-uniform distribution of failure rates across fine-grained error modes. Most failure modes remain in the low-rate regime, but a small subset contributes disproportionately to model differences. In particular, multi-actor coordination errors, visual-consistency errors, and camera-language errors show much larger variation across models than basic acting-related failures. This pattern indicates that the benchmark does not merely rank models by a single aggregate score, but instead exposes model-specific failure profiles at a fine-grained level.

A clear pattern also emerges at the category level. Acting-related failures are generally less severe, whereas multi-actor coordination remains more challenging, especially for action-script alignment and interaction timing. In Atmosphere, lighting and color consistency are among the most persistent error sources across models. The strongest separation appears in Cinematography, where shot progression, 30-degree and 180-degree rule violations, and spatial continuity errors exhibit the widest spread. These results suggest that cinematic grammar and cross-shot continuity remain major bottlenecks for current multi-talker video generation systems.

\subsection{Error Count Distribution}
Figure~\ref{fig:error_count_distribution} shows the distribution of annotated error counts across the three top-level dimensions and their aggregate total. In all three dimensions, the distributions are concentrated in the low-count range. For Acting, 83.58\% of videos contain 0--3 errors, with 1 error being the most common case (25.79\%). For Atmosphere, 87.10\% of videos fall within 0--3 errors, and the mode is 2 errors (26.40\%). For Cinematography, 86.98\% of videos also remain within 0--3 errors, with 0 errors as the largest single bin (24.53\%). High-count tails exist but remain limited: the proportion of videos with 5 or more errors is 6.73\% for Acting, 2.80\% for Atmosphere, and 6.04\% for Cinematography.

The aggregate distribution is broader, which reflects cross-dimensional error co-occurrence. Only 2.31\% of videos contain zero total errors, while the majority of samples (60.14\%) fall into the range of 4--9 total errors. In total, 88.97\% of videos contain no more than 9 errors, and only 11.03\% exceed this level. These results suggest that the benchmark is not dominated by saturated or excessively corrupted examples. Instead, most samples exhibit a limited number of localized failures, while a smaller subset contains denser cross-dimensional error patterns that remain diagnostically informative.

\begin{figure*}[t]
\centering
\includegraphics[width=\textwidth]{figures/error_count_distribution_2x2_700dpi.jpg}
\caption{Distribution of per-video error counts across the three top-level dimensions and the aggregate total.}
\label{fig:error_count_distribution}
\end{figure*}

\subsection{Rationale Consistency Distribution}
\begin{figure}[t]
\centering
\includegraphics[width=\columnwidth]{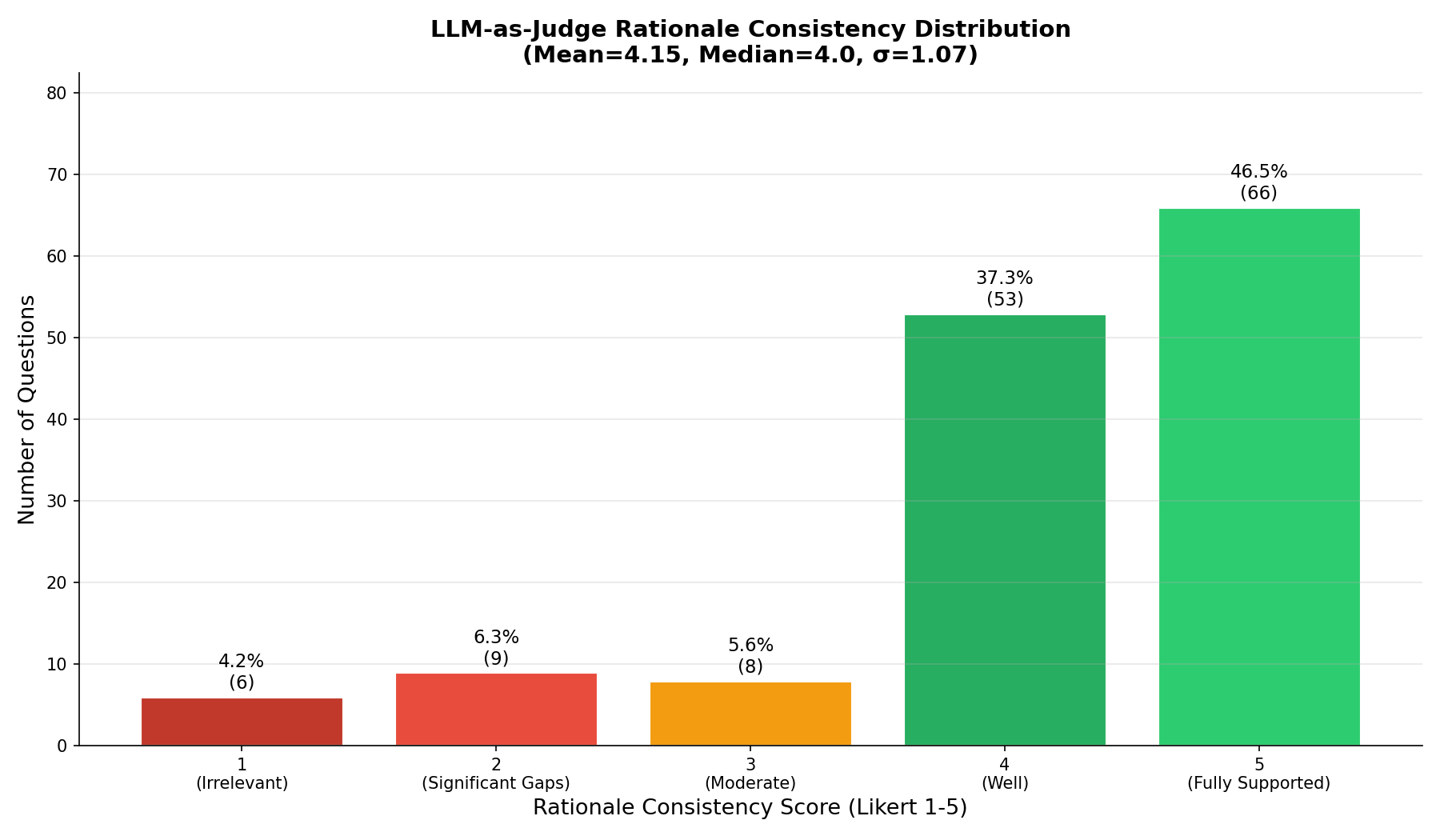}
\caption{Distribution of LLM-as-judge rationale-consistency scores.}
\label{fig:rc_distribution}
\end{figure}

Figure~\ref{fig:rc_distribution} shows the distribution of rationale-consistency scores assigned by the LLM judges. The scores are concentrated in the upper range, with a mean of 4.15 and a median of 4.0. In particular, scores of 4 and 5 account for most samples, indicating that the predicted rationales usually provide clear support for the final selected option. At the same time, the remaining low-score cases show that this metric is not saturated and still captures instances where rationale-answer alignment is weak or incomplete.

\subsection{Failure-Mode Correlation Analysis}

\begin{figure}[t]
\centering
\includegraphics[width=0.8\columnwidth]{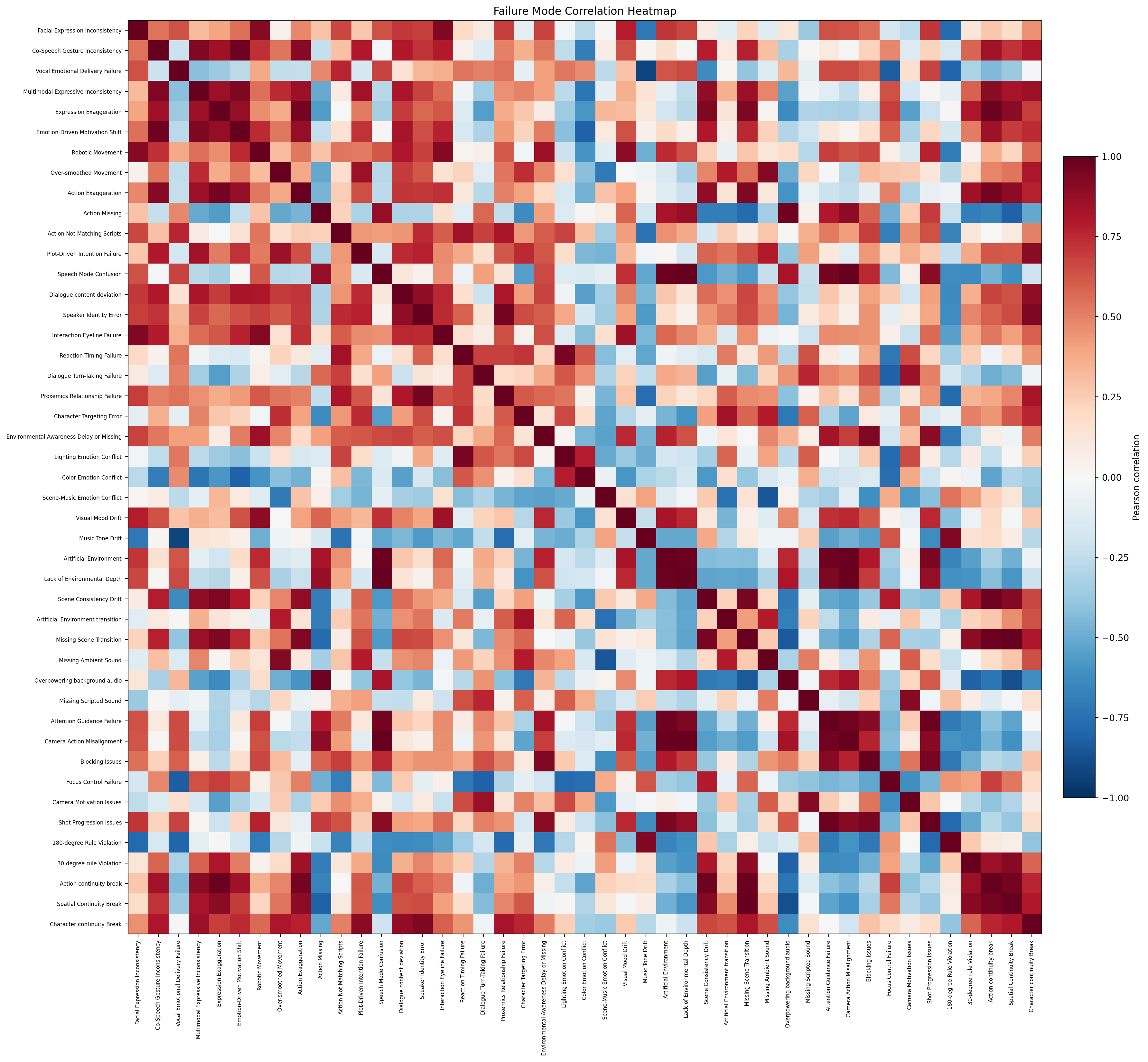}
\caption{Pearson correlation heatmap across fine-grained failure modes.}
\label{fig:failure_mode_correlation}
\end{figure}

Figure~\ref{fig:failure_mode_correlation} shows that fine-grained failure modes are not uniformly independent. Instead, several groups of co-occurring failures emerge as visible blocks of positive correlation. In particular, a cluster of expressive-performance failures can be observed, where co-speech gesture inconsistency, emotion-driven motivation shift, multimodal expressive inconsistency, and exaggerated expression tend to co-occur. A second cluster is centered on continuity and shot grammar, including missing scene transition, scene consistency drift, action continuity break, spatial continuity break, and shot progression issues, which exhibit consistently strong pairwise correlations. At the same time, the matrix also contains weakly correlated or negatively correlated pairs, suggesting that failure propagation is structured rather than global. Overall, these patterns indicate that many generation failures are coupled across neighboring dimensions, especially in multi-scene composition and cinematic continuity.

\section{Failure Cases}
\subsection{Acting-Level Failures}
We present three acting-level examples to illustrate how fine-grained performance failures can emerge in dialogue delivery, body motion, and interaction awareness.

\begin{figure*}[t]
\centering
\includegraphics[width=0.9\textwidth]{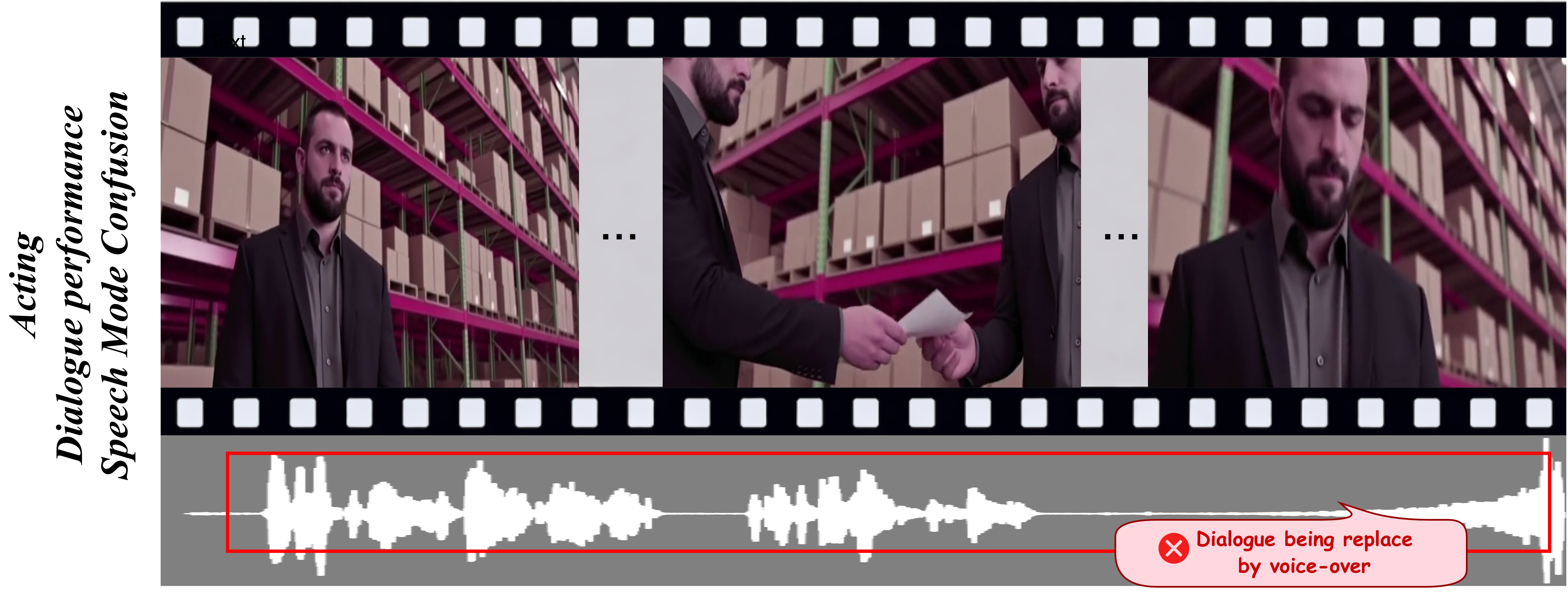}
\caption{Acting-level failure case: speech mode confusion in dialogue performance.}
\label{fig:acting_dialogue_case}
\end{figure*}

\paragraph{Dialogue Performance: Speech Mode Confusion.}
Figure~\ref{fig:acting_dialogue_case} shows a failure case in which the spoken content is realized as detached voice-over rather than synchronized on-screen speech. Although the visual sequence still depicts the character in a dialogue scene, the audio delivery no longer matches the expected speaking mode, creating a mismatch between visible participation and spoken output.

\begin{figure*}[t]
\centering
\includegraphics[width=0.9\textwidth]{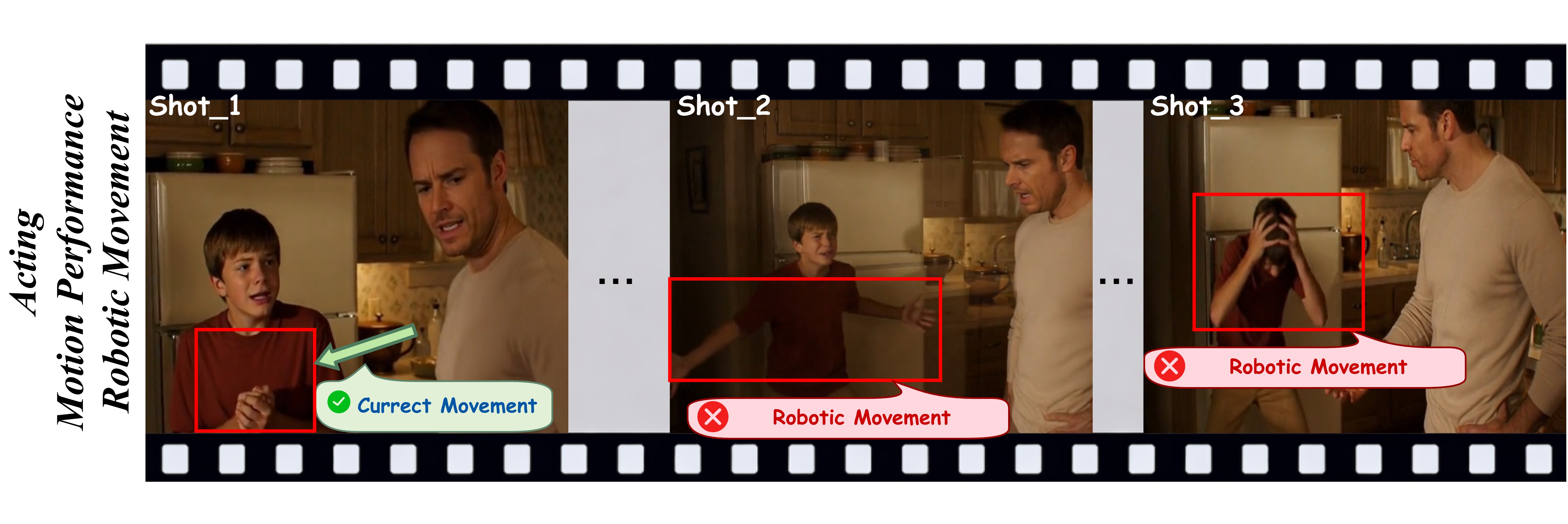}
\caption{Acting-level failure case: robotic movement in motion performance.}
\label{fig:acting_motion_case}
\end{figure*}

\paragraph{Motion Performance: Robotic Movement.}
Figure~\ref{fig:acting_motion_case} presents a case where the character's motion becomes progressively rigid across shots. The initial posture remains plausible, but later frames exhibit stiff and weakly responsive body movement, reducing the natural continuity of the acting performance.

\begin{figure*}[t]
\centering
\includegraphics[width=0.9\textwidth]{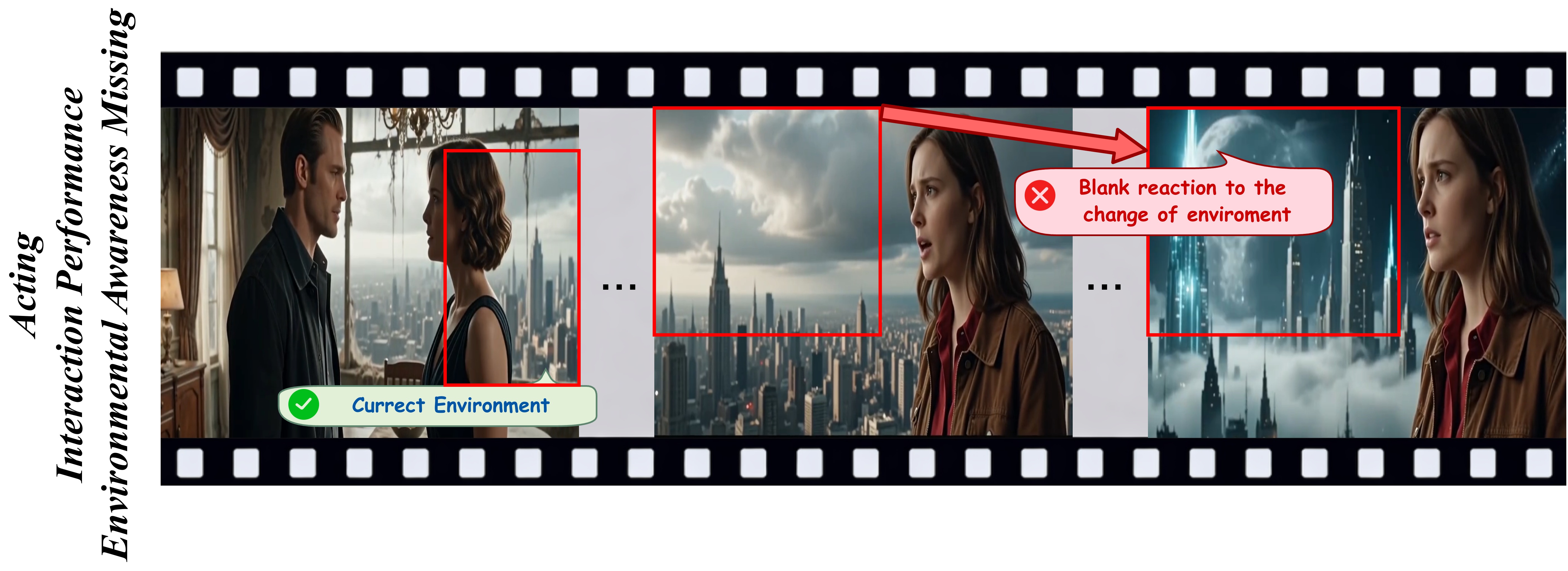}
\caption{Acting-level failure case: missing environmental awareness during interaction.}
\label{fig:acting_interaction_case}
\end{figure*}

\paragraph{Interaction Performance: Environmental Awareness Missing.}
Figure~\ref{fig:acting_interaction_case} highlights an interaction-level failure in which the character does not react appropriately to a salient environmental change. Even though the surrounding scene shifts dramatically, the performance remains largely blank, indicating missing environmental awareness and weakened interaction grounding.

\subsection{Cinematic-Level Failures}
We next present three cinematography-level examples that illustrate failures in intra-shot camera alignment, shot progression, and inter-shot grammar.

\begin{figure*}[t]
\centering
\includegraphics[width=0.9\textwidth]{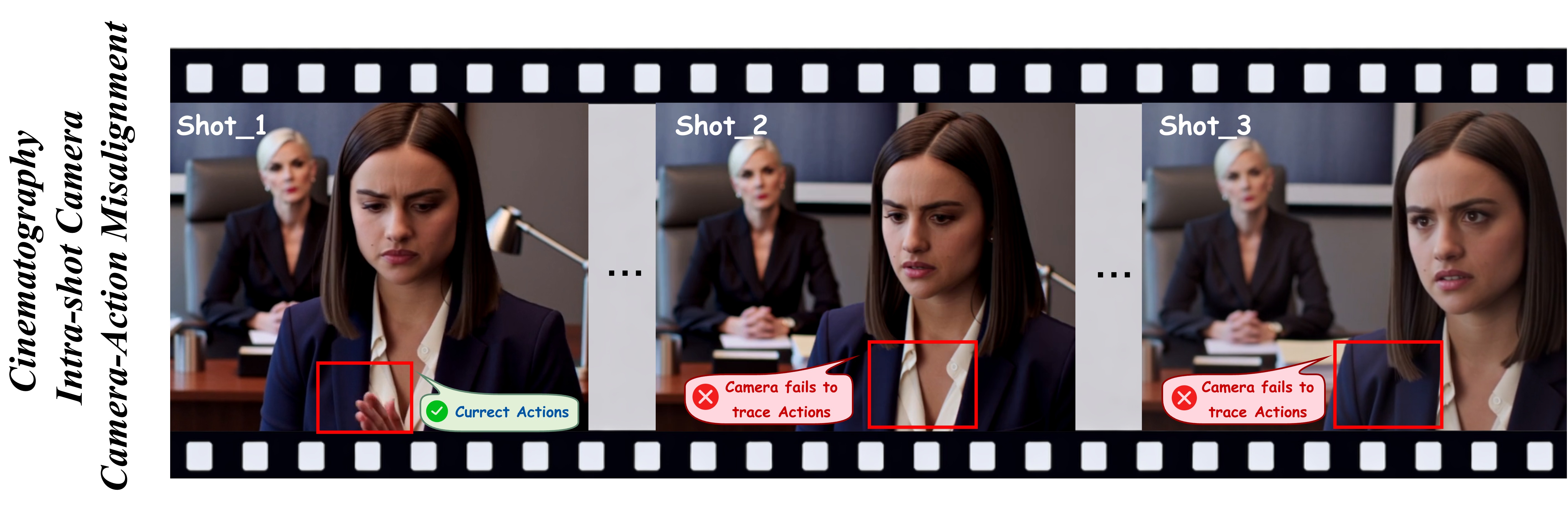}
\caption{Cinematography-level failure case: camera-action misalignment in intra-shot camera control.}
\label{fig:cine_camera_action_case}
\end{figure*}

\paragraph{Intra-shot Camera: Camera-Action Misalignment.}
Figure~\ref{fig:cine_camera_action_case} shows a case where the camera fails to track the action-bearing region across consecutive shots. Although the character remains visible, the framing no longer follows the hand motion that should be emphasized, resulting in weak camera-action alignment and reduced guidance of viewer attention.

\begin{figure*}[t]
\centering
\includegraphics[width=0.9\textwidth]{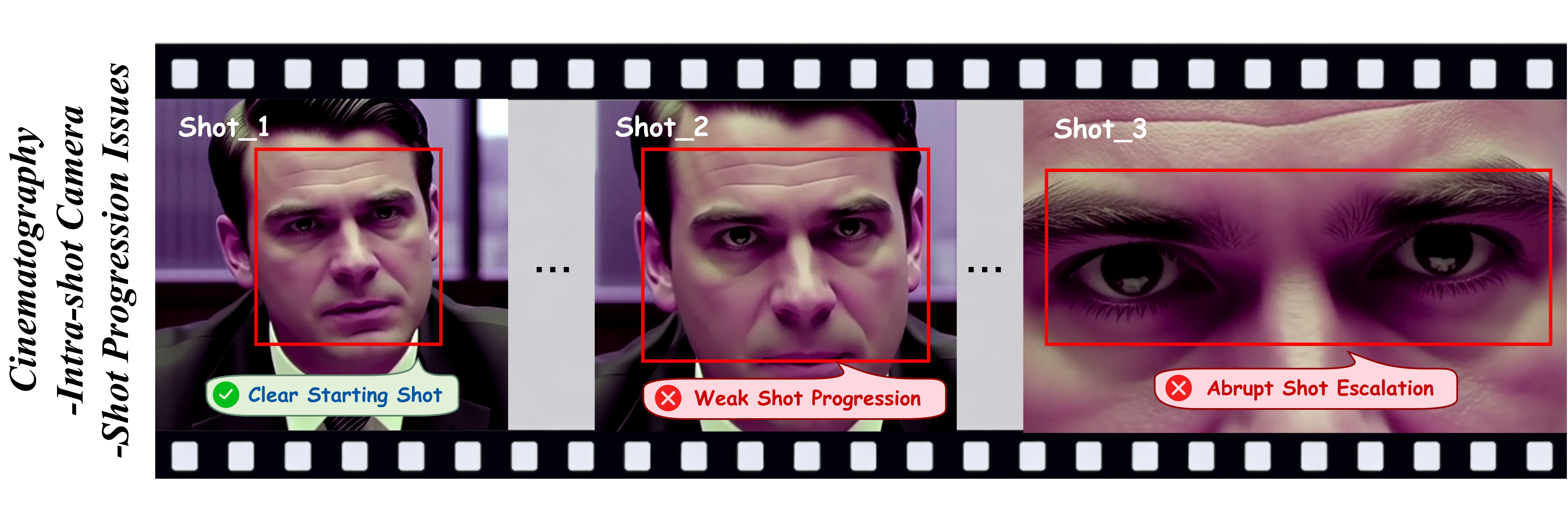}
\caption{Cinematography-level failure case: weak shot progression and abrupt escalation.}
\label{fig:cine_shot_progression_case}
\end{figure*}

\paragraph{Inter-shot Camera: Shot Progression Issues.}
Figure~\ref{fig:cine_shot_progression_case} presents an example of weak shot progression. The sequence begins with a readable base shot, but the subsequent escalation becomes abrupt and visually disproportionate, weakening the intended step-by-step progression of shot intensity.

\begin{figure*}[t]
\centering
\includegraphics[width=0.9\textwidth]{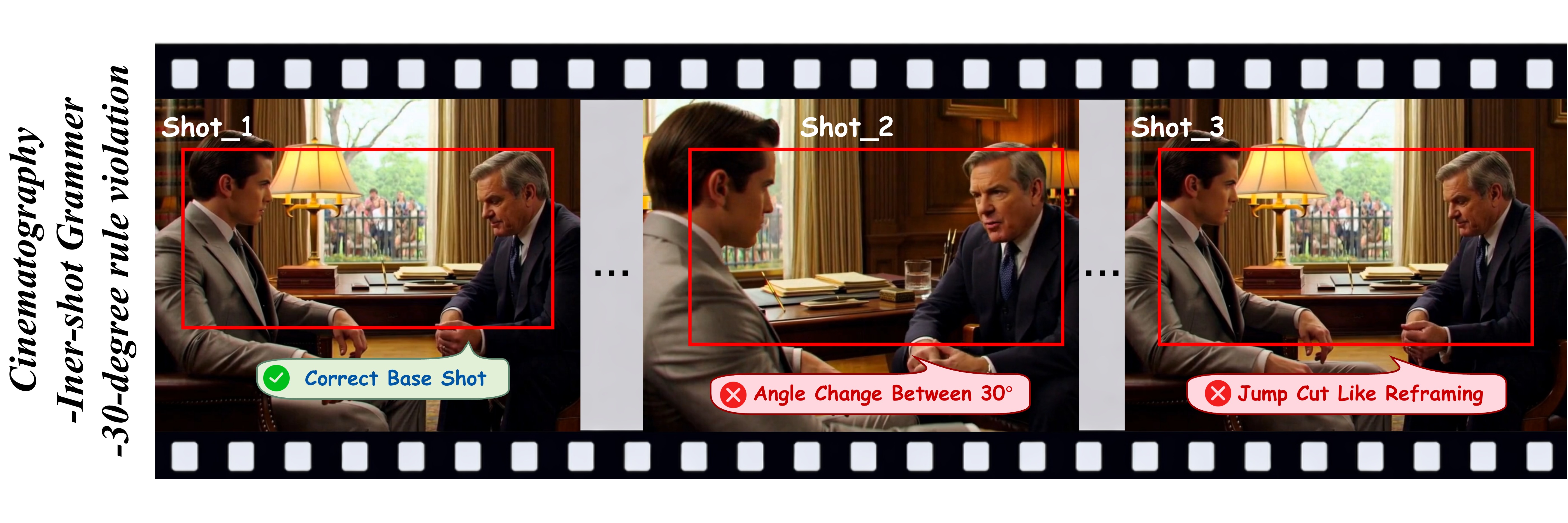}
\caption{Cinematography-level failure case: 30-degree rule violation in inter-shot grammar.}
\label{fig:cine_30degree_case}
\end{figure*}

\paragraph{Inter-shot Grammar: 30-Degree Rule Violation.}
Figure~\ref{fig:cine_30degree_case} illustrates a violation of inter-shot grammar. After a correct base shot, the camera changes by less than a sufficient angular margin, which creates a jump-cut-like reframing instead of a perceptually clean shot transition. This makes the cut feel unstable even though the scene content itself remains similar.

\subsection{Atmosphere-Level Failures}
We further present an atmosphere-level example to illustrate how failures in mood construction can arise from inconsistency between audiovisual cues and scene emotion.

\begin{figure*}[t]
\centering
\includegraphics[width=0.9\textwidth]{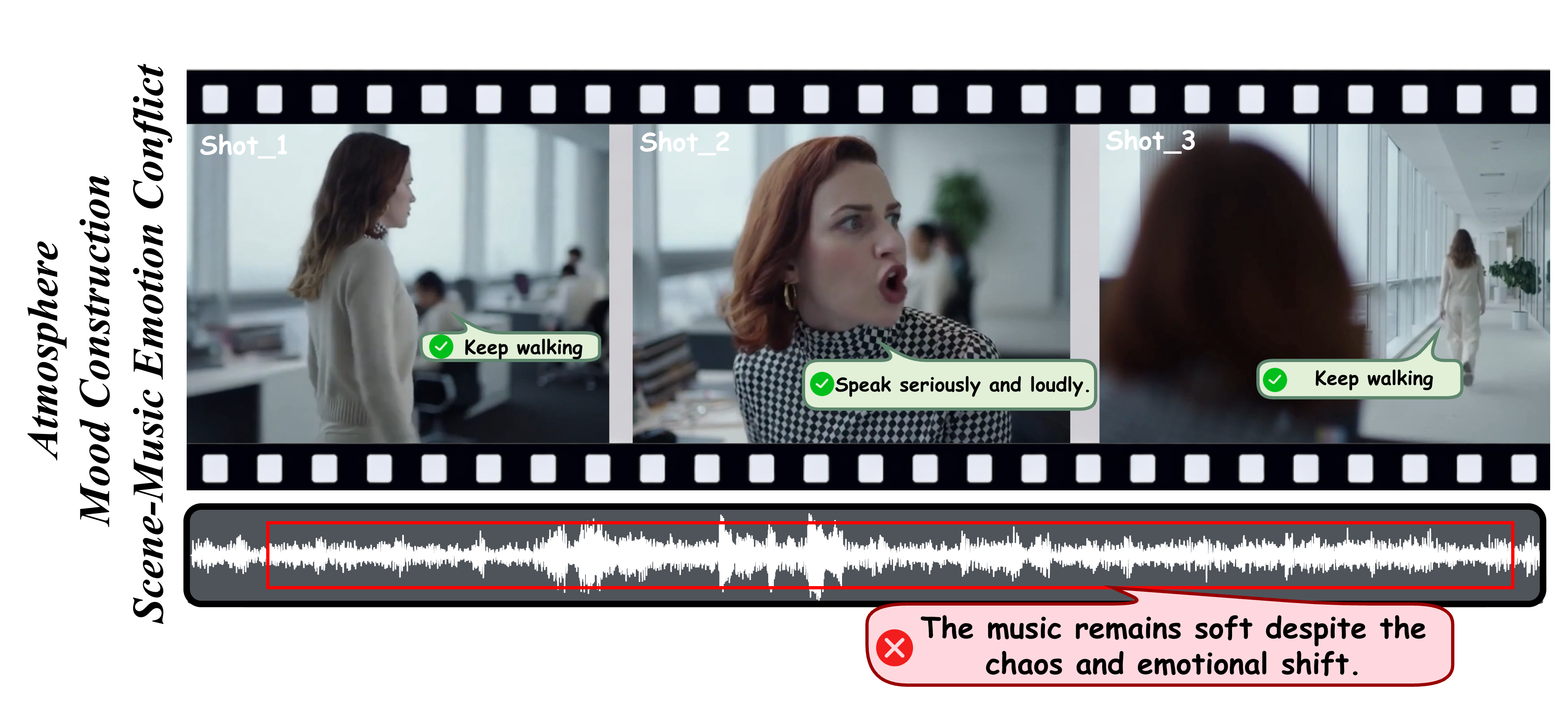}
\caption{Atmosphere-level failure case: background music remains soft despite the scene's escalating chaos and emotional conflict.}
\label{fig:atmosphere}
\end{figure*}

\paragraph{Mood Construction: Scene--Music Emotion Conflict.}
Figure~\ref{fig:atmosphere} shows a case where the background music remains soft and gentle even as the scene escalates into visible chaos and emotional confrontation. Although the character actions and shot progression indicate rising tension, the soundtrack does not adapt accordingly, producing a mismatch between the constructed atmosphere and the scene's emotional trajectory.

\section{Human Evaluation Details}
\subsection{Evaluation Setup}
Human validation is embedded in benchmark construction through an expert-pool protocol, rather than conducted as a separate post-hoc user study. Each review is conducted on a scene-level short video. For each QA item, two experts are first sampled from a pool of 22 human verifiers for independent review. Their task is to decide whether the proposed question-answer pair is consistent with the visible failure evidence, the target failure mode, and the fixed annotation framework.

\subsection{Annotation Interface}
\begin{figure*}[t]
\centering
\includegraphics[width=0.9\textwidth]{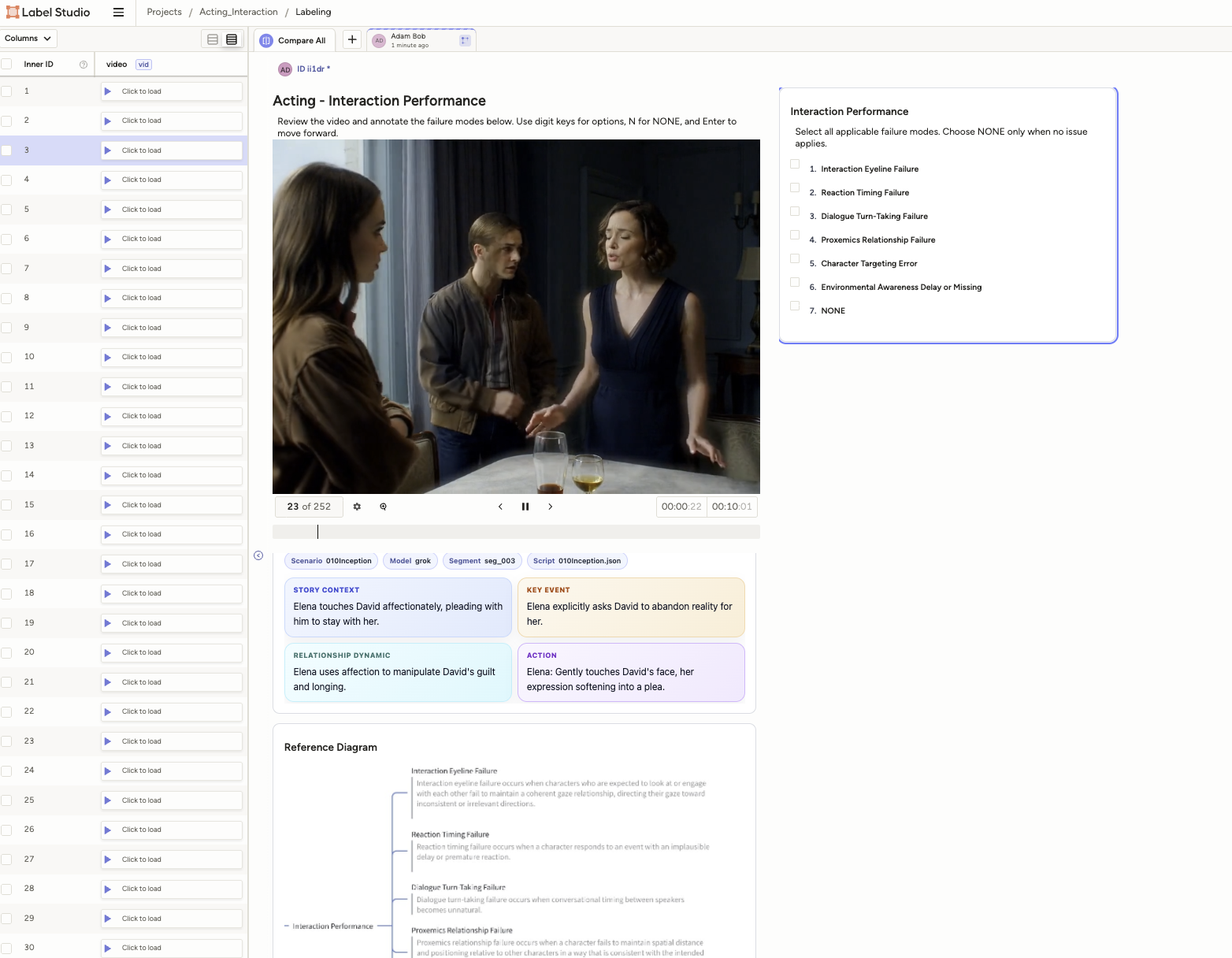}
\caption{Customized Label Studio Interface for MTAVG-Bench 2.0 Annotation}
\label{fig:annotation_interface}
\end{figure*}

Figure~\ref{fig:annotation_interface} shows the customized Label Studio interface used for benchmark refinement. Reviewers are presented with a scene-level clip, candidate failure-mode options, and concise explanatory fields such as Story Context, Key Event, Relationship Dynamic, and Action. The interface also provides dimension-specific guidance and reference cues to support consistent annotation. In interaction-oriented cases, these cues help reviewers judge whether character actions and responses are properly aligned with the intended target and the surrounding context.

\section{Ethical Considerations}
This benchmark is designed to evaluate and diagnose cinematic-level generation failures, rather than to make value judgments about real individuals or social groups. The scripts used in the benchmark do not involve personal information from real people, and the evaluation process does not depend on identity-specific information about real actors.

Methodologically, our setup resembles the practice of ``imitation shooting'' in film education: cinematic scenes are reconstructed for analysis and evaluation. Concretely, we first generate an initial clip using a text-to-audio-video (T2VA) model and then continue the generation through the I2VA pipeline. The goal is to assess and diagnose generated cinematic content, not to reproduce real individuals or to infer attributes about them.

\paragraph{Copyright Notice.}
The benchmark is intended for cinematic understanding and evaluation based on classic film material. It is not intended for redistributing original movie content or exploiting the likeness of real performers. Because some benchmark materials are inspired by or derived from existing films, copyright and fair-use boundaries should be carefully considered when using, releasing, or redistributing related materials.

\ifdefined\appendixasinput
\else
\end{document}
\fi

\end{document}